%% file: emnlp.tex
\documentclass[11pt]{article}

\usepackage{acl}

\usepackage{times}
\usepackage{latexsym}
\usepackage[T1]{fontenc}
\usepackage[utf8]{inputenc}
\usepackage{microtype}
\makeatletter
\providecommand{\MT@suspend@tagging}{}
\providecommand{\MT@resume@tagging}{}
\makeatother
\usepackage{graphicx}
\usepackage{amsmath}
\usepackage{amssymb}
\usepackage{booktabs}
\usepackage{xcolor}
\usepackage{enumitem}
\usepackage{subcaption}
\IfFileExists{trimclip.sty}{%
  \usepackage{trimclip}%
  \newcommand{\shortrightarrow}{\clipbox*{{.25\width} 0pt {\width} {\height}} \textrightarrow}%
}{%
  \newcommand{\shortrightarrow}{\rightarrow}%
}

\newcommand{\reals}{\mathbb{R}}

\newcommand{\email}[1]{{\fontsize{11pt}{11pt}\selectfont\texttt{#1}}}

\title{Better World Models Can Lead to Better Post-Training Performance}

\author{
Prakhar Gupta \\
University of Michigan \\
\email{prakharg@umich.edu} \\
\And
Henry Conklin \\
Princeton University \\
\email{henry.conklin@princeton.edu} \\
\AND
Sarah-Jane Leslie \\
Princeton University \\
\email{sjleslie@princeton.edu} \\
\And
Andrew Lee \\
Harvard University \\
\email{andrewlee@g.harvard.edu} \\
}

\begin{document}
\maketitle

% ============================================================================
\begin{abstract}
We study how explicit world-modeling objectives affect the internal representations and downstream capability of Transformers, using Rubik's Cubes as our training domain.
We ask:
(1) how does explicitly pretraining a world model affect a model's latent representations,
(2) how does world-model quality affect post-training performance, and
(3) how should a finite data budget be split between pretraining and task-specific fine-tuning?
We compare standard action-prediction fine-tuning with two strategies that add explicit state-prediction supervision: state pretraining followed by fine-tuning, and joint action and state training. We measure task accuracy after Group Relative Policy Optimization (GRPO).
We further find that explicit world-modeling yields better representations in terms of higher probing accuracy and steerability of the model, and that better representations yield larger gains from GRPO, especially on harder cube states.
Finally, when the fine-tuning budget is held fixed, probe accuracy strongly predicts the GRPO improvement. Under a fixed total data budget, accuracy is maximized by allocating only a small fraction to pretraining.
\end{abstract}

\input{sections/1_intro}
\input{sections/2_setup}

\input{sections/3_pretraining}

\input{sections/4_grpo}
\input{sections/5_data_allocation}
\input{sections/6_related_work}
\input{sections/7_conclusion}

% ============================================================================
% Suppress line numbers in the back matter (limitations, ethics, references,
% appendix) to avoid the close-gutter spacing where right-column line numbers
% visually crowd the right-column text. Main body keeps its line numbers.
\nolinenumbers

\section*{Limitations}

We note a few limitations.

\textbf{(i)} Per-distance accuracy values at the largest 2$\times$2 distances rest on small test-set counts (394 at distance 10, 4 at distance 11). We exclude distance 11 from quantitative analyses and treat distance-10 numbers as indicative; aggregate claims are robust because most test mass sits at distances 8--9.

\textbf{(ii)} The 3$\times$3 data-allocation experiments (Section~\ref{sec:data_allocation}) restrict evaluation to the $\leq 11$ distance subset for cross-size comparability with 2$\times$2; whether the same allocation optimum persists when harder cubes ($d{>}11$) are included is an open question. The exact best split may also depend on model size and training setup.

\textbf{(iii)} Beyond linear probes and causal interventions, additional measures of how much the model relies on its cube-state representations (e.g., mutual information between predictions and probe-decoded states) could provide complementary evidence; we leave such analyses to future work.

\textbf{(iv)} Both cube sizes belong to one synthetic task family with deterministic dynamics, known intermediate states, and exactly measurable success. Extending these findings to stochastic environments, additional planning domains, or natural-language tasks is important future work.

% ============================================================================
\section*{Ethics Statement}

This work uses entirely synthetic data generated from deterministic Rubik's Cube mechanics, with no human subjects, personal data, or web-scraped corpora involved. The trained Transformers are small (8--16 layers) and solve a controlled planning task; we are not aware of foreseeable downstream applications that raise ethical concerns beyond those already attached to research on representation learning and reinforcement-learning post-training generally.

% ============================================================================
\bibliography{references}

% ============================================================================
\appendix

\section{Hyperparameters}
\label{sec:appx_hyperparameters}

\subsection{Model Architecture}

Model architecture details for the 2$\times$2 and 3$\times$3 models are listed in Table~\ref{tab:appx_arch}.

\begin{table}[!ht]
\centering
\small
\begin{tabular}{lcc}
\toprule
\textbf{Parameter} & \textbf{2$\times$2} & \textbf{3$\times$3} \\
\midrule
Layers ($L$)            & 8   & 16 \\
Attention heads         & 8   & 16 \\
Hidden dimension ($d$)  & 512 & 1024 \\
Vocabulary size         & 16  & 25 \\
Max sequence length     & 50  & 80 \\
Squares ($P$)           & 24  & 54 \\
\bottomrule
\end{tabular}
\caption{\label{tab:appx_arch}GPT-2--style architecture per cube size~\citep{radford2019language}.}
\end{table}

\subsection{Fine-Tuning, Pretraining, Joint Training}

Hyperparameters for fine-tuning, pretraining, and joint training are listed in Table~\ref{tab:appx_hyper}.

\begin{table}[!ht]
\centering
\small
\begin{tabular}{lcc}
\toprule
\textbf{Parameter} & \textbf{2$\times$2} & \textbf{3$\times$3} \\
\midrule
Learning rate     & 1e-5  & 1e-4 \\
Weight decay      & 0.01  & 0.01 \\
Batch size        & 64    & 128 \\
Optimizer         & AdamW & AdamW \\
\bottomrule
\end{tabular}
\caption{\label{tab:appx_hyper}Hyperparameters for fine-tuning, pretraining, and joint training. The core nine-configuration experiments use early-stopping patience 10; the data-allocation experiments (Section~\ref{sec:data_allocation}) save end-of-epoch checkpoints with early stopping disabled, so that the number of epochs is a controlled variable.}
\end{table}

\subsection{GRPO}

GRPO hyperparameters are listed in Table~\ref{tab:appx_grpo_hyper}.

\begin{table}[!ht]
\centering
\small
\begin{tabular}{lcc}
\toprule
\textbf{Parameter} & \textbf{2$\times$2} & \textbf{3$\times$3} \\
\midrule
Learning rate                  & 1e-5  & 1e-5 \\
Per-device train batch size    & 256   & 256 \\
Per-device eval batch size     & 128   & 128 \\
Number of rollouts             & 8     & 8 \\
KL penalty ($\beta$)           & 0.01  & 0.01 \\
Weight decay                   & 0.01  & 0.01 \\
Max prompt length              & 24    & 54 \\
Max completion (core / alloc.) & 13 / 13 & 26 / 14 \\
Epochs (core / alloc.)         & 3 / 1 & 3 / 1 \\
Loss type                      & DR-GRPO & DR-GRPO \\
Optimizer                      & AdamW & AdamW \\
\bottomrule
\end{tabular}
\caption{\label{tab:appx_grpo_hyper}GRPO hyperparameters. ``Core'' refers to the nine-configuration experiments of Section~\ref{sec:planning}; ``alloc.'' refers to the data-allocation experiments of Section~\ref{sec:data_allocation}. For 3$\times$3 data-allocation, the test set is restricted to distance $\leq 11$, so the max completion length is set to $11{+}3{=}14$.}
\end{table}

\subsection{Probing}

Probe-training hyperparameters are listed in Table~\ref{tab:appx_probe_hyper}.

\begin{table}[!ht]
\centering
\small
\begin{tabular}{ll}
\toprule
\textbf{Parameter}    & \textbf{Value} \\
\midrule
Learning rate         & 1e-3 \\
Weight decay          & 0.01 \\
Batch size            & 32 \\
Early-stopping patience & 10 \\
Epochs                & 1 \\
Optimizer             & AdamW \\
\bottomrule
\end{tabular}
\caption{\label{tab:appx_probe_hyper}Linear probe training hyperparameters (identical across cube sizes).}
\end{table}

% ============================================================================
\section{Data Details}
\label{sec:appx_data}

\paragraph{2$\times$2 test set.}
Table~\ref{tab:appx_test_222} gives the per-distance breakdown.
Distances 8--9 hold ${\sim}85\%$ of the test examples; distances 10--11 are extremely sparse (only 4 examples at distance 11), so per-distance claims at distance 11 are excluded from quantitative analyses.

\begin{table}[!ht]
\centering
\small
\begin{tabular}{c|rr}
\toprule
\textbf{Distance} & \textbf{Count} & \textbf{\%} \\
\midrule
3  & 20      & 0.02\% \\
4  & 112     & 0.10\% \\
5  & 608     & 0.53\% \\
6  & 3{,}232 & 2.82\% \\
7  & 13{,}256 & 11.57\% \\
8  & 45{,}088 & 39.34\% \\
9  & 51{,}892 & 45.28\% \\
10 & 394     & 0.34\% \\
11 & 4       & 0.003\% \\
\midrule
\textbf{Total} & \textbf{114{,}606} & \textbf{100\%} \\
\bottomrule
\end{tabular}
\caption{\label{tab:appx_test_222}2$\times$2 test-set breakdown by optimal distance.}
\end{table}

\paragraph{3$\times$3 train/test disjointness.}
The uniform 3$\times$3 training set is built from the cube20.org distance-20 corpus and all of its optimal sub-paths; after deduplication by scramble state we obtain $16{,}556{,}726$ unique (scramble, solution) pairs, which we cap at $222{,}116$ per distance (distances 1--4 are below the cap and kept in full), yielding $3{,}600{,}000$ training states.
Train/test disjointness is verified by enumerating $363{,}735$ intermediate states at distance $>3$ along each test trajectory and confirming zero overlap with any training entry's solution path.

% ============================================================================
\section{Intervention Procedure Details}
\label{sec:appx_interventions}

The intervention setup is described in Section~\ref{sec:rq1_pretrain}; we note two implementation details here.
\textbf{Pair construction.}
For 2$\times$2 we enumerate all ${\sim}3.67{\times}10^6$ reachable states with their optimal solutions and use this lookup table to score whether an intervened-on move is good for the target state $\mathcal{T}$.
For 3$\times$3 the $\sim 4.3{\times}10^{19}$-state space precludes enumeration, so we use an on-the-fly Kociemba-style solver to score moves.
\textbf{Margin sweep.}
The adaptive margin $m$ was tuned per cube size. We swept $m \in \{0.5, 1, 2\}$ for 2$\times$2 and $m \in \{0.5, 1, 2, 4\}$ for 3$\times$3, selecting the value with the highest mean intervention success across the three training strategies: $m{=}1$ for 2$\times$2, $m{=}4$ for 3$\times$3.

% ============================================================================
\section{Data-Allocation Experiment Designs}
\label{sec:appx_sweep_design}

\subsection{Pretraining-Amount (Section~\ref{subsec:fixed_sft})}
We split $D_1$ in half: a fixed SFT slice (always used in full, 3 SFT epochs) and a pretrain pool.
The pretrain pool is sub-sampled at nine fractions $k/8$ for $k \in \{0, 1, \dots, 8\}$, and pretraining runs for $N \in \{1, \dots, 5\}$ epochs independently of SFT.
This yields $1 + 8 \times 5 = 41$ SFT checkpoints per cube size and 41 corresponding SFT+GRPO checkpoints.
Pretraining is purely additive: adding more pretraining never displaces SFT data, so the SFT data is held fixed across configurations.

\subsection{Budget-Split (Section~\ref{subsec:ratio_sweep})}
We split $D_1$ at nine equidistant ratios $k/8$ for $k \in \{0, \dots, 8\}$, with the pretrain portion holding $k/8$ of $D_1$ and the SFT portion holding $(8-k)/8$.
For each ratio we train at $N \in \{1, \dots, 5\}$ epochs, with pretraining and SFT each running for $N$ epochs (epoch-matched).
GRPO is 1 epoch on $D_2$.
$k{=}0$ is the SFT-only baseline; $k{=}8$ is pretrain-only and receives no GRPO. Total: 40 SFT + 40 SFT+GRPO checkpoints per cube size.

% ============================================================================
\section{Additional Results}
\label{sec:appx_results}

\subsection{Distributional Steering}
\label{subsec:appx_intervene_prob_mass}
\begin{figure*}[t]
\centering
\begin{minipage}[t]{0.47\textwidth}\centering
  {\footnotesize\textbf{\textsf{2$\times$2$\times$2}}}\\[2pt]
  \includegraphics[width=\textwidth]{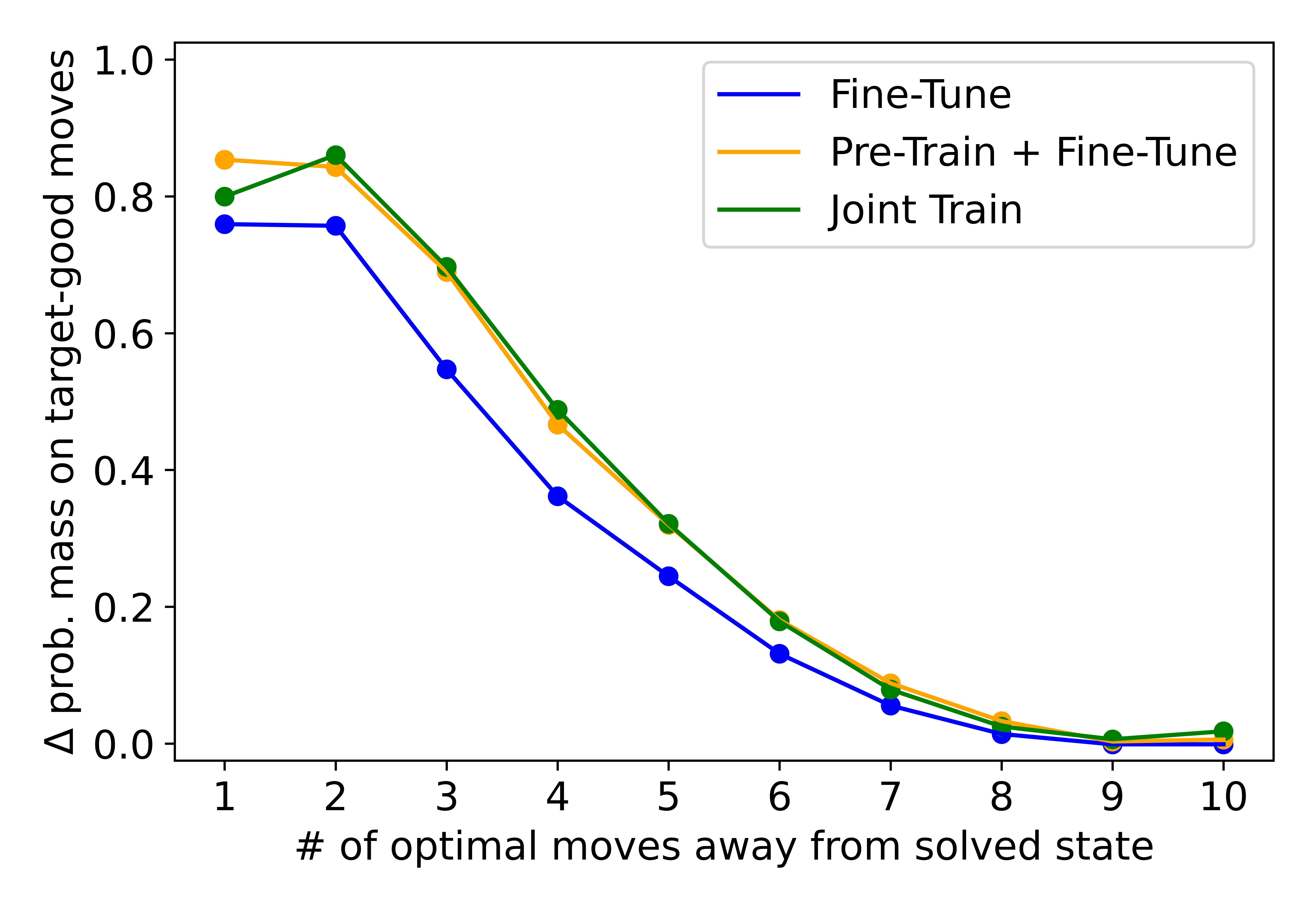}
\end{minipage}\hfill
\begin{minipage}[t]{0.47\textwidth}\centering
  {\footnotesize\textbf{\textsf{3$\times$3$\times$3}}}\\[2pt]
  \includegraphics[width=\textwidth]{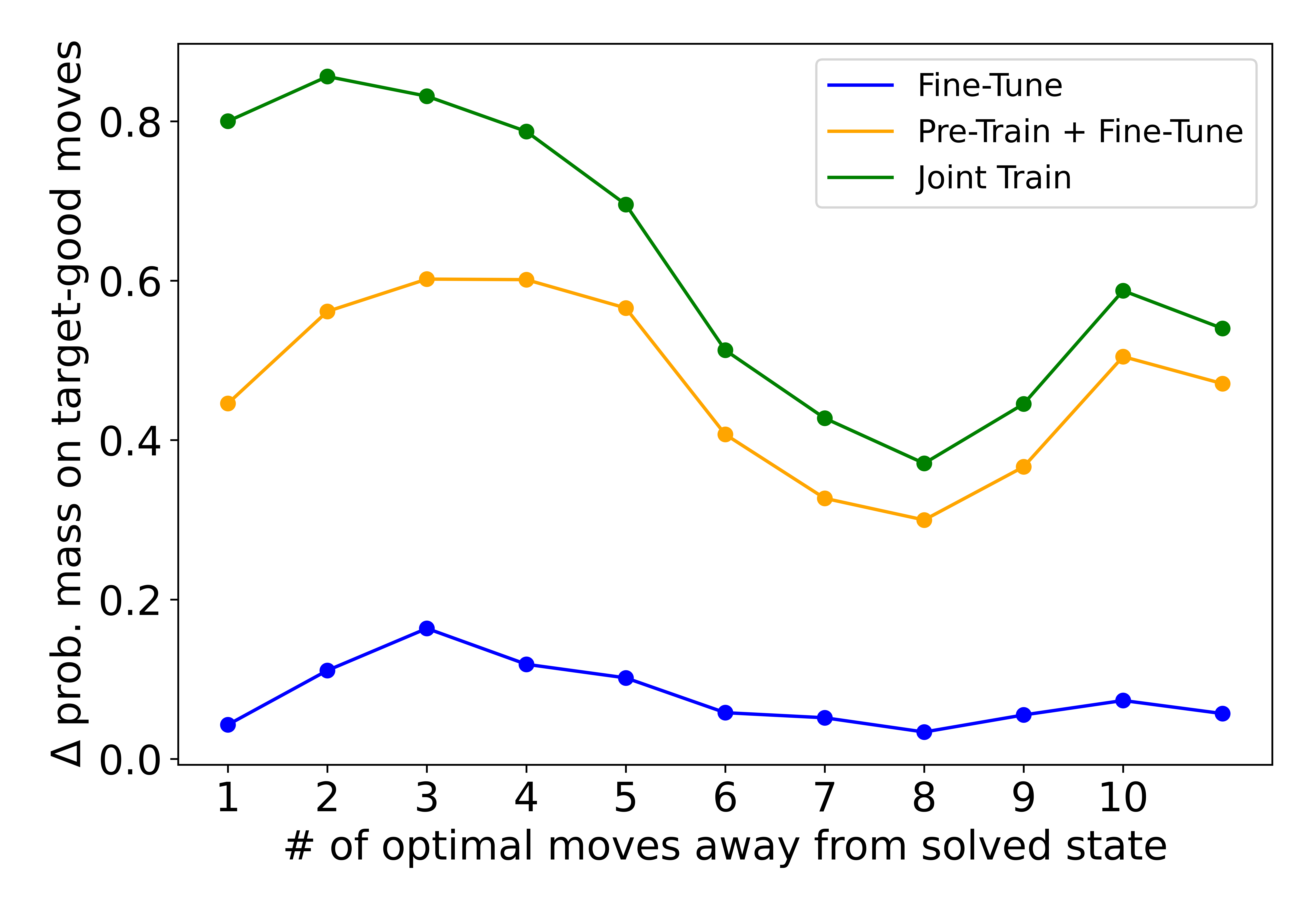}
\end{minipage}
\caption{\label{fig:intervene_prob_mass}\textbf{Distributional steering.}
Change in probability mass on target-good moves (post-intervention minus pre-intervention), broken down by cube distance.}
\end{figure*}

Figure~\ref{fig:intervene_prob_mass} is the distributional-steering counterpart of Figure~\ref{fig:intervene_results}: rather than the top-1 predicted move, we report the change in probability mass on the set of target-good moves (post-intervention minus pre-intervention). The qualitative picture matches Figure~\ref{fig:intervene_results} on both cube sizes: models trained with explicit state-prediction supervision shift more mass toward the target-good moves.

\subsection{3$\times$3 Nine-Configuration Comparison}

Table~\ref{tab:appx_grpo_333} gives the full 3$\times$3 task-accuracy numbers for the nine training configurations of Section~\ref{subsec:grpo_accuracy}, with means and standard deviations over 5 seeds.

For the five 2$\times$2 $D_1+\mathrm{GRPO}(D_2)$ seeds, exact-optimal
accuracy was $8.0\pm2.8\%$, $18.8\pm1.6\%$, and $10.4\pm1.3\%$ for
FT, Pretrain+FT, and Joint, respectively. Average extra moves were
$2.34\pm0.14$, $1.72\pm0.08$, and $2.27\pm0.12$.

\begin{table}[!ht]
\centering
\small
\setlength{\tabcolsep}{4pt}
\begin{tabular}{lccc}
\toprule
\textbf{Strategy} & $D_1$ & $D_1{\cup}D_2$ & $D_1${+}GRPO \\
\midrule
Fine-Tune     & $14.71{\pm}0.36$ & $15.03{\pm}0.61$ & $16.92{\pm}0.38$ \\
Pre-Train+FT  & $23.50{\pm}1.21$ & $24.20{\pm}1.42$ & $27.94{\pm}0.96$ \\
Joint Train   & $25.69{\pm}0.34$ & $26.78{\pm}0.93$ & $29.55{\pm}0.39$ \\
\bottomrule
\end{tabular}
\caption{\label{tab:appx_grpo_333}3$\times$3 task accuracy (\%), mean $\pm$ std over 5 seeds. $D_1${+}GRPO uses 3-epoch GRPO.}
\end{table}

\subsection{Pretraining-Amount Heatmaps}
\label{subsec:appx_fxsft_heatmaps}

Tables~\ref{tab:appx_fxsft_sft_222}--\ref{tab:appx_fxsft_grpo_333} give the per-cell SFT-only and SFT+GRPO accuracies for the pretraining-amount experiment on each cube. The corresponding per-cell GRPO gain is the difference of the two tables.

\begin{table}[!ht]
\centering
\small
\begin{tabular}{c|ccccc}
\toprule
\textbf{Frac.} & \textbf{1ep} & \textbf{2ep} & \textbf{3ep} & \textbf{4ep} & \textbf{5ep} \\
\midrule
1/8 & 0.1 & 0.6 & 1.4 & 3.2 & 5.3 \\
2/8 & 0.6 & 2.0 & 4.7 & 6.8 & 8.0 \\
3/8 & 1.2 & 4.8 & 7.7 & 9.3 & 10.0 \\
4/8 & 2.0 & 7.4 & 9.3 & 10.3 & 10.8 \\
5/8 & 4.6 & 9.2 & 10.7 & 11.5 & 12.0 \\
6/8 & 5.5 & 9.6 & 11.1 & 11.5 & 12.2 \\
7/8 & 6.5 & 10.4 & 11.6 & 11.8 & 12.3 \\
8/8 & 7.7 & 11.0 & 12.1 & 12.5 & \textbf{12.9} \\
\bottomrule
\end{tabular}
\caption{\label{tab:appx_fxsft_sft_222}2$\times$2 SFT-only accuracy (\%). The $k{=}0$ no-pretrain baseline is $0.01\%$ (effectively $0$).}
\end{table}

\begin{table}[!ht]
\centering
\small
\begin{tabular}{c|ccccc}
\toprule
\textbf{Frac.} & \textbf{1ep} & \textbf{2ep} & \textbf{3ep} & \textbf{4ep} & \textbf{5ep} \\
\midrule
1/8 & 0.1 & 0.8 & 2.6 & 10.0 & 17.9 \\
2/8 & 0.7 & 3.7 & 15.5 & 22.2 & 26.0 \\
3/8 & 2.3 & 15.1 & 24.9 & 28.8 & 30.2 \\
4/8 & 3.9 & 24.0 & 29.1 & 31.5 & 32.8 \\
5/8 & 15.5 & 30.1 & 33.4 & 34.6 & 35.8 \\
6/8 & 18.3 & 29.6 & 33.1 & 34.8 & 35.4 \\
7/8 & 20.4 & 31.5 & 33.9 & 35.2 & 36.2 \\
8/8 & 24.6 & 33.6 & 35.8 & 36.9 & \textbf{37.7} \\
\bottomrule
\end{tabular}
\caption{\label{tab:appx_fxsft_grpo_222}2$\times$2 SFT+GRPO accuracy (\%). The $k{=}0$ no-pretrain baseline is $0.01\%$.}
\end{table}

\begin{table}[!ht]
\centering
\small
\begin{tabular}{c|ccccc}
\toprule
\textbf{Frac.} & \textbf{1ep} & \textbf{2ep} & \textbf{3ep} & \textbf{4ep} & \textbf{5ep} \\
\midrule
1/8 & 14.1 & 20.2 & 23.6 & 24.9 & 25.9 \\
2/8 & 21.4 & 26.7 & 29.8 & 34.0 & 35.5 \\
3/8 & 22.1 & 27.3 & 32.1 & 34.5 & 35.7 \\
4/8 & 24.0 & 28.6 & 32.6 & 34.6 & 35.7 \\
5/8 & 25.4 & 32.6 & 35.5 & 36.3 & 36.4 \\
6/8 & 28.0 & 33.9 & 36.3 & 37.1 & 37.6 \\
7/8 & 26.8 & 31.7 & 35.1 & 36.8 & 37.4 \\
8/8 & 33.1 & 36.5 & 37.7 & 37.9 & \textbf{38.3} \\
\bottomrule
\end{tabular}
\caption{\label{tab:appx_fxsft_sft_333}3$\times$3 SFT-only accuracy (\%) on the $\leq 11$ test subset. The $k{=}0$ no-pretrain baseline is $12.11\%$.}
\end{table}

\begin{table}[!ht]
\centering
\small
\begin{tabular}{c|ccccc}
\toprule
\textbf{Frac.} & \textbf{1ep} & \textbf{2ep} & \textbf{3ep} & \textbf{4ep} & \textbf{5ep} \\
\midrule
1/8 & 20.3 & 25.7 & 28.9 & 29.9 & 30.9 \\
2/8 & 27.0 & 31.5 & 35.0 & 39.2 & 41.3 \\
3/8 & 27.7 & 31.8 & 37.0 & 39.8 & 40.9 \\
4/8 & 28.6 & 32.8 & 37.1 & 39.9 & 41.5 \\
5/8 & 30.2 & 37.6 & 40.5 & 42.0 & 42.4 \\
6/8 & 32.5 & 39.2 & 42.2 & 42.8 & 43.3 \\
7/8 & 31.1 & 36.8 & 40.2 & 42.6 & 43.0 \\
8/8 & 38.2 & 42.6 & 43.7 & 44.2 & \textbf{44.4} \\
\bottomrule
\end{tabular}
\caption{\label{tab:appx_fxsft_grpo_333}3$\times$3 SFT+GRPO accuracy (\%) on the $\leq 11$ test subset. The $k{=}0$ no-pretrain baseline is $18.51\%$.}
\end{table}

\subsection{Per-Distance Accuracy at Peak Budget-Split (2$\times$2)}

Table~\ref{tab:appx_per_dist} gives the per-distance accuracy at the peak 2$\times$2 budget-split configuration ($k{=}1/8$, $N{=}5$). GRPO's gains are concentrated at distances 8--10, where the base model is far from both ceiling and floor.

\begin{table}[!ht]
\centering
\small
\begin{tabular}{crrrr}
\toprule
\textbf{Dist.} & \textbf{N} & \textbf{SFT} & \textbf{+GRPO} & \textbf{$\Delta$} \\
\midrule
3  & 20      & 100.0 & 100.0 & (ceiling) \\
4  & 112     & 100.0 & 100.0 & (ceiling) \\
5  & 608     & 97.4  & 94.7  & $-2.6$ \\
6  & 3{,}232 & 85.6  & 89.2  & $+3.7$ \\
7  & 13{,}256 & 61.2 & 77.2  & $+16.0$ \\
8  & 45{,}088 & 35.1 & 67.3  & $+32.2$ \\
9  & 51{,}892 & 15.8 & 63.8  & $\mathbf{+48.0}$ \\
10 & 394     & 1.8   & 48.5  & $+46.7$ \\
\bottomrule
\end{tabular}
\caption{\label{tab:appx_per_dist}Per-distance accuracy (\%) at the peak 2$\times$2 budget-split configuration ($k{=}1/8$, $N{=}5$). Distance 11 ($N{=}4$) is omitted as statistically meaningless.}
\end{table}

\subsection{Probe Accuracy vs.\ GRPO Gain}
\label{subsec:appx_bins}

Table~\ref{tab:probe_grpo_correlation} summarizes the correlations behind Figure~\ref{fig:probe_grpo_scatter}; Table~\ref{tab:appx_bins} gives the per-bin mean GRPO gain $\pm$ SEM.

\begin{table}[!ht]
\centering
\small
\begin{tabular}{lcc}
\toprule
\textbf{Configurations} & \textbf{Pearson} $r$ & \textbf{Spearman} $\rho$ \\
\midrule
2$\times$2, all 41           & $+0.80$ & $+0.96$ \\
2$\times$2, above floor       & $\mathbf{+0.94}$ & $+0.94$ \\
3$\times$3, all 41           & $+0.35$ & $+0.45$ \\
3$\times$3, above floor       & $\mathbf{+0.84}$ & $+0.87$ \\
\bottomrule
\end{tabular}
\caption{\label{tab:probe_grpo_correlation}Probe accuracy versus GRPO gain when only the amount of pretraining is varied ($N{=}41$ per cube). The above-floor rows drop the lowest equal-count probe-accuracy bin ($N{=}34$).}
\end{table}

\begin{table}[!ht]
\centering
\small
\begin{tabular}{ccc}
\toprule
\textbf{Probe-accuracy bin} & \textbf{2$\times$2} & \textbf{3$\times$3} \\
\midrule
1 (floor) & $0.6{\pm}0.3$  & $5.6{\pm}0.2$ \\
2         & $9.4{\pm}1.5$  & $4.7{\pm}0.1$ \\
3         & $17.5{\pm}0.9$ & $4.9{\pm}0.1$ \\
4         & $20.6{\pm}0.8$ & $5.2{\pm}0.2$ \\
5         & $22.3{\pm}0.5$ & $5.7{\pm}0.1$ \\
6         & $23.9{\pm}0.3$ & $6.0{\pm}0.1$ \\
\bottomrule
\end{tabular}
\caption{\label{tab:appx_bins}Mean GRPO gain (percentage points) $\pm$ SEM within each equal-count probe-accuracy bin (about 7 models each, $N{=}6$ in the top bin), ordered from lowest to highest world-model quality.}
\end{table}

\paragraph{Floor-bin diagnostics.}
The two floor bins fail for opposite reasons, visible in their within-bin correlations.
On 2$\times$2 the floor bin (probe accuracy $0.39$--$0.70$) collects models with post-SFT accuracy near zero ($0$--$2\%$), leaving GRPO no reward to amplify; within this bin, gain rises with SFT accuracy at Pearson $+0.97$, so the bin mean ($0.6$pp) is dragged down by the SFT-near-$0\%$ models.
On 3$\times$3 the floor bin (probe accuracy $0.56$--$0.62$) instead collects low-base-accuracy models (SFT accuracy $12$--$24\%$), and gain correlates \emph{negatively} with both SFT accuracy (Pearson $-0.91$) and probe accuracy (Pearson $-0.84$) within the bin. This is a headroom effect: from such a low base, raw-percentage gains track available room to improve more than they track world-model quality, which is why the bin mean ($5.6$pp) sits above the next two bins.
Above the floor both cubes show a monotone climb (Table~\ref{tab:probe_grpo_correlation}); probe accuracy additionally tracks absolute task accuracy nearly rank-perfectly on 3$\times$3 (Pearson $r{=}0.96$ for post-SFT and $r{=}0.98$ for post-GRPO accuracy, Spearman $\rho{=}0.99$ for both).

\end{document}

%% file: sections/1_intro.tex
\section{Introduction}
\label{sec:introduction}

Language models demonstrate impressive reasoning capabilities.
These models typically go through multiple training stages, including pretraining on generic data, fine-tuning on task-specific data, and post-training using reinforcement learning for further improvements~\citep{ouyang2022training, guo2025deepseek}.
However, it is unclear how each stage affects the internal representations of the model, and in turn how such representations affect each stage.

We study these questions in a controlled setting.
Namely, we train Transformers on a task that requires planning: solving Rubik's Cubes of two sizes, 2$\times$2$\times$2 and 3$\times$3$\times$3.
Not only does the task require planning, but it also requires the model to learn a ``world model'', by which the model must understand a latent cube state (i.e., ``world''), and how its predictions (actions) affect the cube state.

Researchers have demonstrated that Transformers can implicitly learn world models when trained via next-token prediction.
For example, a model trained on transcripts of \emph{moves} being played in a board game such as Othello can learn to model the latent board-state, despite never being given any priors regarding the board~\citep{li2023emergent, nanda2023emergent}. %, and similar results have been shown for chess~\citep{karvonen2024emergent}.
While these works show the emergence of a world model, they do not study its relationship with downstream training stages nor capabilities.
Thus in this work we ask three questions:

\begin{enumerate}[noitemsep,topsep=2pt,leftmargin=2.2em]
    \item[\textbf{RQ1}] How does explicitly pretraining a world model affect the representations of the model?
    \item[\textbf{RQ2}] How does the quality of world models affect the model's accuracy post-training?
    \item[\textbf{RQ3}] How should a finite training data budget be allocated between world-model pretraining and task-specific fine-tuning?
\end{enumerate}

From our experiments we find the following.
\textbf{(1)} Explicitly pretraining a world model leads to more robust representations, in terms of higher probe accuracies as well as better steerability of the model, suggesting a higher reliance of the model on its latent cube state representations.
\textbf{(2)} Models with better world models achieve higher planning accuracy after reinforcement learning, and probe accuracy strongly predicts the GRPO improvement when the fine-tuning budget is held fixed, a direct quantitative link between representation quality and post-training outcome.
\textbf{(3)} When pretraining and fine-tuning compete for a fixed data budget, accuracy is maximized with a small fraction of pretraining (roughly an eighth) in the long-training regime, and this optimum shifts toward more pretraining when training is short.
Together, our findings demonstrate the close relationship between the model's representations and its effects on downstream training stages.
%The probing result in finding (1) and the GRPO-gain result in finding (2) replicate on both 2$\times$2$\times$2 and 3$\times$3$\times$3, and the optimal data-allocation point in finding (3) is identical across the two cube sizes, strengthening the evidence for generality.\footnote{Causal steering has so far been measured on 2$\times$2$\times$2 only.}

%% file: sections/2_setup.tex
\section{Setup, Notation, and Data}
\label{sec:setups}

We study Transformer models trained to solve Rubik's Cubes of two sizes: 2$\times$2$\times$2, which consists of 6 faces and 4 squares per face (24 squares total), and 3$\times$3$\times$3, which consists of 6 faces and 9 squares per face (54 squares total).
We henceforth refer to each cube as 2$\times$2 or 3$\times$3.
We formulate our task as sequence modeling with the following data: $\mathcal{D} = \{(\mathcal{S}_i, \mathcal{A}_i)\}_{i=1}^N$ where $\mathcal{S}$ is a scrambled cube state representing the state of all $P$ squares, $\mathcal{S} := [x_1, \dots x_{P}]$, and $\mathcal{A}$ is the \emph{optimal} sequence of moves that solves the cube, $\mathcal{A} := [m_1, \dots, m_n]$, where $n$ is the cube distance for the initial state $\mathcal{S}$ from being solved.
Each state token $x$ specifies a square's color ($x \in \mathcal{C}$, $\vert\mathcal{C}\vert = 6$).
Each move token $m$ specifies an action ($m \in \mathcal{M}$, with $\vert\mathcal{M}\vert = 9$ for 2$\times$2 and $\vert\mathcal{M}\vert = 18$ for 3$\times$3 in Singmaster's notation\footnote{Singmaster notation labels each $90^\circ$ face turn by the initial of the face (U, D, L, R, F, B), with a prime denoting a counter-clockwise turn and a 2 denoting a $180^\circ$ turn. The 2$\times$2$\times$2 fixes one corner and uses three faces, giving $3 \times 3 = 9$ moves; the 3$\times$3$\times$3 uses all six, giving $6 \times 3 = 18$.}).
Importantly, intermediate cube states from applying moves $m$ are not part of the token sequence, and must be implicitly learned by the model. In the state-supervised settings, they are used only as auxiliary classification targets.
We notate the hidden states of the model $\mathbf{h}^\ell$ and token unembeddings $\mathbf{U}$.

\subsection{Training Setups}
\label{subsec:training_setups}

\paragraph{Standard Fine-Tuning.}
In our standard setting, we train on $\mathcal{D}$ with the standard next-token prediction objective.
The model is given as input a scrambled cube $\mathcal{S}$ and is trained to auto-regressively at every remaining timestep to predict the sequence $\mathcal{A}$:
$\mathcal{L}_{FT} = \text{CrossEntropy}(\mathbf{U}\mathbf{h}^{L-1}_t, \mathcal{A}_t)$.

\paragraph{Pretraining a World Model.}
In some settings, we add an optional pretraining step to explicitly learn a world model before standard fine-tuning.
We use a different dataset, $\{(\mathcal{S}_j, \mathcal{\hat{A}}_j)\}^N$, where $\mathcal{\hat{A}}$ is now a sequence of \emph{random} moves.
The goal of pretraining is to explicitly predict the latent cube state (rather than next moves) given a sequence of moves.
This is done by substituting the unembedding layer $\mathbf{U}$ with $P$ alternative classification heads $\mathbf{W}_i \in \reals^{|\mathcal{C}| \times d}$, one for each square $i$, that each classify the correct state of the $P$ squares of the cube, with the following loss:
\begin{equation*}
\mathcal{L}_{PT} = \frac{1}{P} \sum_{i=0}^{P-1} \text{CrossEntropy}(\mathbf{W}_i \mathbf{h}^{L-1}_t, \mathcal{S}^*_{t,i})
\end{equation*}
where $\mathbf{W}_i$ is the classifier for the $i$-th square and $\mathcal{S}^*_{t, i}$ is the groundtruth color for the $i$-th square after applying the first $t$ moves in $\mathcal{\hat{A}}$ to the initial state $\mathcal{S}$.
Once pretrained, the model is fine-tuned to predict optimal moves, re-using all the weights except for those of $\mathbf{W}$.

\paragraph{Joint Training.}
An alternative approach is to learn the two objectives jointly in a single stage.
Here the model uses both classification heads ($\mathbf{U}, \mathbf{W}$) with loss $\mathcal{L}_{joint} = \mathcal{L}_{FT} + \alpha \mathcal{L}_{PT}$.
We use $\alpha = 1.0$ in all joint-training experiments. Joint Training otherwise uses the same data and setup as FT, with the state-prediction loss added on the same trajectories.

\paragraph{Post-Training: GRPO.}
Once a model is trained with one of the recipes above, we study the effects of post-training with reinforcement learning, using GRPO~\citep{shao2024deepseekmath}.
Given a training sample (a scrambled cube state), GRPO samples multiple rollouts and assigns a reward per rollout (1 for rollouts that solve the cube, 0 otherwise), which is used to compute a loss term.
Hyperparameters for all training stages are in Appendix~\ref{sec:appx_hyperparameters}.

\begin{table}[t]
\centering
\small
\begin{tabular}{lcc}
\toprule
 & \textbf{2$\times$2$\times$2} & \textbf{3$\times$3$\times$3} \\
\midrule
Training states & 3,559,554 & 3,600,000 \\
Test states & 114,606 & 100,000 \\
Max cube distance & 11 & 20 \\
Optimal solutions? & Yes & Yes \\
Squares ($P$) & 24 & 54 \\
Moves ($|\mathcal{M}|$) & 9 & 18 \\
\bottomrule
\end{tabular}
\caption{\textbf{Summary of our two task data.}}
\label{tab:data_comparison}
\end{table}

\begin{table}[t]
\centering
\small
\setlength{\tabcolsep}{2pt}
\begin{tabular}{ll}
\toprule
\textbf{Description} & \textbf{Training Setup} \\
\midrule
FT & $\mathcal{L}_{FT}(D_1 \cup D_2)$ \\
PT + FT & $\mathcal{L}_{PT}(\mathcal{D}_\text{pre}) \shortrightarrow \mathcal{L}_{FT}(D_1 \cup D_2)$ \\
Joint Train & $\mathcal{L}_{joint}(D_1 \cup D_2)$ \\
\midrule
FT + GRPO & $\mathcal{L}_{FT}(D_1) \shortrightarrow \mathcal{L}_{GRPO}(D_2)$ \\
PT + FT + GRPO & $\scriptsize{\mathcal{L}_{PT}(\mathcal{D}_\text{pre})}
\shortrightarrow \scriptsize{\mathcal{L}_{FT}(D_1)}
\shortrightarrow \scriptsize{\mathcal{L}_{GRPO}(D_2)}$ \\
Joint Train + GRPO & $\mathcal{L}_{joint}(D_1) \shortrightarrow \mathcal{L}_{GRPO}(D_2)$ \\
\bottomrule
\end{tabular}
\caption{\textbf{Summary of training configurations.}}
\label{tab:training_setups}
\end{table}

\begin{figure*}[t]
\centering
\includegraphics[width=0.94\textwidth]{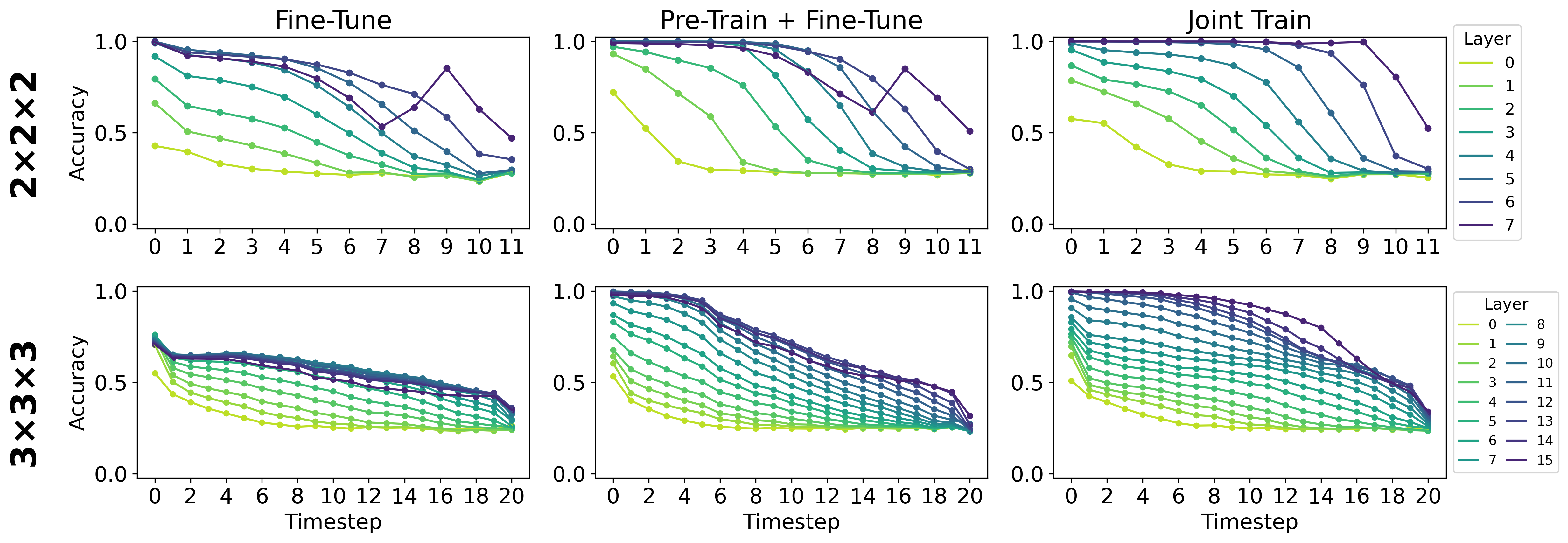}
\caption{\label{fig:probe_accuracy_222}\textbf{Probing accuracy, mean over 5 seeds.}
For both cubes, models trained with explicit state-prediction supervision have higher cube-state probe accuracy than models trained only on move prediction.}
\end{figure*}
\subsection{Data}
\label{subsec:data}

We design our data splits for controlled comparisons across training strategies.
Both datasets contain \emph{optimal} solutions: every solution is a shortest path to the solved state.

\paragraph{2$\times$2 Data.}
A 2$\times$2 cube has roughly $3.67 \times 10^6$ possible states, all enumerable via breadth-first search.
We carefully split between training and test sets to minimize overlap in intermediate cube states.
To do so, we build the test set by constructing scrambled states maximally far from being solved (11 moves away), applying optimal moves toward the solved state, and collecting all intermediate states from these paths.
This yields $114{,}606$ test states.
All other trajectories that do not cross these intermediate states are used as training data, ensuring the intermediate cube states in the train and test sets are disjoint.\footnote{This disjointness does not hold for cube states fewer than two moves from being solved. The number of such states is small.}
This yields $3{,}559{,}554$ training states.
The training data is further split into two equal halves $D_1, D_2$, each $1{,}779{,}777$ states.

\paragraph{3$\times$3 Data.}
A 3$\times$3 cube has approximately $4.3 \times 10^{19}$ possible states, making exhaustive enumeration infeasible.
We start from the ``God's Number'' project~\citep{rokicki2014diameter}, which provides $\sim$1.13M states that are exactly 20 moves from solved in the half-turn metric.
Since each state's 20-move solution is optimal, any suffix of that solution is also an optimal solution for the corresponding intermediate state.
For each 20-move solution $S_1 \dots S_{20}$, we generate sub-paths by applying $S_1 \dots S_k$ for $k \in \{1, \dots, 19\}$, yielding new scrambles at distances 19 down to 1.
After deduplication by scramble state, we cap each distance at a target count, producing a uniformly distributed training set of $3{,}600{,}000$ states and a disjoint test set of $100{,}000$ states (disjointness verified, see Appendix~\ref{sec:appx_data}).
As with 2$\times$2$\times$2, the training data is split into two equal halves $D_1, D_2$. GRPO is always applied to $D_2$.
Table~\ref{tab:data_comparison} summarizes both datasets.

\subsection{Training Configurations}
\label{subsec:training_configs}

Table~\ref{tab:training_setups} summarizes our training setups.
The first three rows are designed to answer RQ1, the latter three to study RQ2. The data-allocation experiments in Section~\ref{sec:data_allocation} add further configurations for RQ3.
For the nine core configurations, we train 5 runs with different seeds per cube size.
For 2$\times$2, we train 8-layer Transformers with 8 attention heads and dimension 512. For 3$\times$3, we scale up to 16 layers, 16 heads, and dimension 1024. We use the two cube sizes to test whether the same qualitative patterns repeat, rather than to directly compare task difficulty. %, reflecting the larger action space and longer solutions.
Full hyperparameters are in Appendix~\ref{sec:appx_hyperparameters}.

%% file: sections/3_pretraining.tex
\section{RQ1: Effect of Pretraining on World Model Representations}
\label{sec:rq1_pretrain}

\paragraph{Probing.}
How does explicitly training a world model affect the model's representations?
We answer this by (1) training linear probes to decode cube state information, and (2) steering the model's predictions using the linear probes.
%We follow prior work~\citep{nanda2023emergent} to train linear probes.
To train our probes, at each timestep $t$ and layer $\ell$, we learn $P$ linear decoders (one for each square $i$), $\mathbf{V}_{t, i}^\ell \in \reals^{6 \times d}$, that minimize
$\mathcal{L}_{probe}(\mathbf{h}^\ell_t, i) = \text{CrossEntropy}(\mathbf{V}_{t, i}^\ell \mathbf{h}^\ell_t, \mathcal{S}_{t,i})$.
The projection $\mathbf{V}_{t, i}\mathbf{h}^\ell_t \in \reals^{6}$ gives logits for the 6 possible colors for square $i$.

Figure~\ref{fig:probe_accuracy_222} shows the results, the top row for the 2$\times$2 cube and the bottom for the 3$\times$3.
For both cube setups, explicitly pretraining a world model leads to an improved cube state representation over standard fine-tuning.

\paragraph{Steering.}
Next we study how dependent each model's predictions are on its internal cube state representations using causal interventions.
To do so, we use the linear probes to override the model's representation of the current cube state $\mathcal{S}$ with that of an alternative target state $\mathcal{T}$.
A successful intervention means the model's new prediction reflects a good move for $\mathcal{T}$ instead of $\mathcal{S}$, with a higher success rate suggesting higher dependency on the cube state representations.

We construct the intervention dataset from states at distances $d \in \{3, \dots, 11\}$ in the 2$\times$2 test set.
For each state we append an optimal move sequence to the input context. For every (distance, timestep) combination $(d, t)$ with $t \in \{0, \dots, d-1\}$ along the optimal solution path, we sample up to $1{,}000$ random instances, yielding $36{,}440$ samples in aggregate.
For each instance with groundtruth state $\mathcal{S}$, we randomly select an alternative state $\mathcal{T}$ at the same optimal distance, ensuring $\mathcal{S}$ and $\mathcal{T}$ share no common good next moves.
We intervene to remove information about all $P$ squares of $\mathcal{S}$ and add information about $\mathcal{T}$.
Let the groundtruth colors of $\mathcal{S}$ be $[x_0, \dots x_{P-1}]$, let $\mathbf{v}_i = \mathbf{V}^\ell_{t, i}[x_i] \in \reals^d$ be the probe vector for color $x_i$ of square $i$, let $[y_0, \dots y_{P-1}]$ be the colors of $\mathcal{T}$, and $\tilde{\mathbf{v}}_i = \mathbf{V}^\ell_{t, i}[y_i]$.
The intervention is
\begin{equation*}
\mathbf{h}^{\ell}_t =
\Big(
    \mathbf{I}-\!\sum_{i=0}^{P-1}\!\frac{
        \mathbf{v}_i\mathbf{v}_i^\top
    }{
        ||\mathbf{v}_i||^2
    }\Big)
\mathbf{h}^\ell_t
+ \sum_{i=0}^{P-1}\alpha_i\tilde{\mathbf{v}}_i,
\end{equation*}
where the first term projects out $\mathcal{S}$ and the second adds $\mathcal{T}$.
After intervening, $\mathbf{h}^\ell_t$ is re-normalized to match its original norm.
For the 2$\times$2 model, which has 8 layers, we intervene on the 5th through 7th blocks, leaving the final block untouched. For the 3$\times$3 model, which has 16 layers, we intervene on the 9th through 15th blocks, leaving the final block untouched.
We \emph{adaptively} select each $\alpha_i$ such that $(\mathbf{V}^\ell_{t, i}\mathbf{h}^\ell_t)_{y_i} \geq \max_{c\neq y_i}(\mathbf{V}^\ell_{t,i}\mathbf{h}^\ell_t)_c + m$, scaling $\alpha_i$ up until the target color is encoded with a margin $m$.
We sweep $m$ on both cubes and find $m=1$ works best for 2$\times$2 and $m=4$ for 3$\times$3.

\begin{figure*}[t]
\centering
\begin{minipage}[t]{0.47\textwidth}\centering
  {\footnotesize\textbf{\textsf{2$\times$2$\times$2}}}\\[2pt]
  \includegraphics[width=\textwidth]{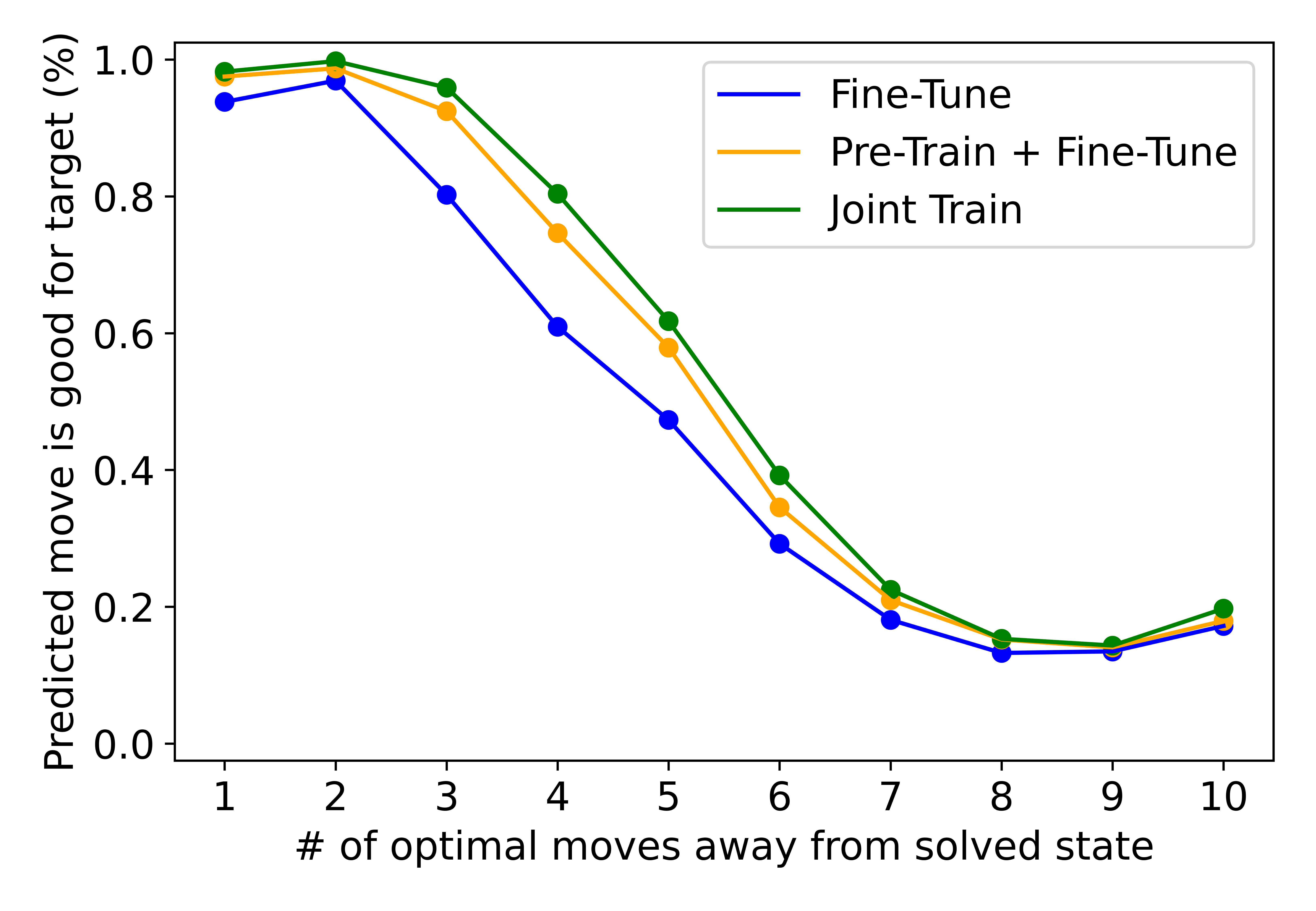}
\end{minipage}\hfill
\begin{minipage}[t]{0.47\textwidth}\centering
  {\footnotesize\textbf{\textsf{3$\times$3$\times$3}}}\\[2pt]
  \includegraphics[width=\textwidth]{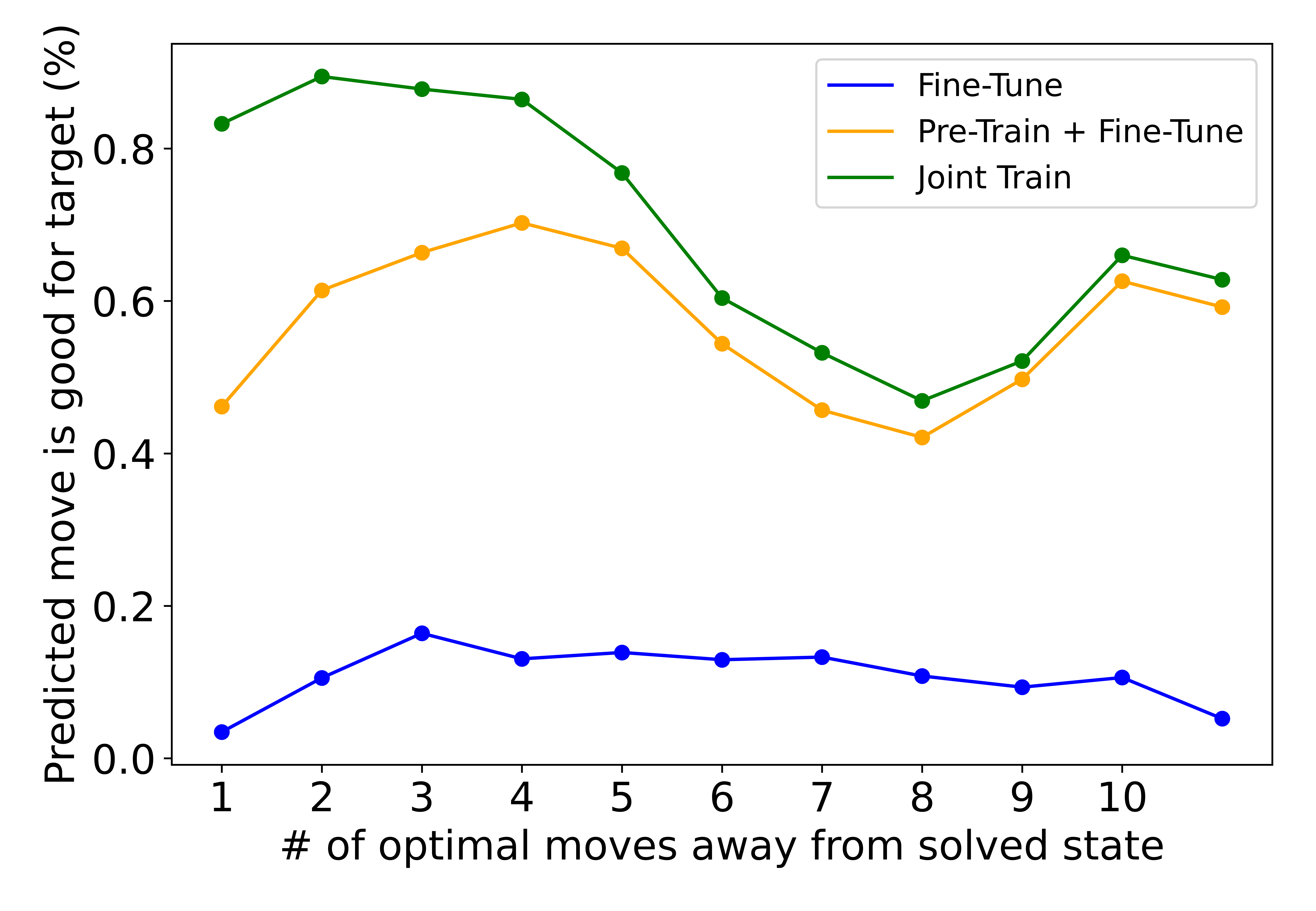}
\end{minipage}
\caption{\label{fig:intervene_results}\textbf{Intervention success rates.}
\emph{Left:} 2$\times$2. \emph{Right:} 3$\times$3 ($\leq 11$ subset).
Each panel reports the fraction of interventions where the model predicts a good move for the target state $\mathcal{T}$ instead of the original state $\mathcal{S}$, broken down by cube distance.
Models trained with explicit state-prediction supervision have higher intervention success rates on both cube sizes.
The distributional-steering counterpart (change in probability mass on target-good moves) is reported in Appendix~\ref{sec:appx_results}, Figure~\ref{fig:intervene_prob_mass}.}
\end{figure*}

Figure~\ref{fig:intervene_results} reports the fraction of interventions where the model's top predicted move is a good move for $\mathcal{T}$ instead of $\mathcal{S}$, broken down by cube distance and shown separately for each cube size.
The pretrained and joint-trained models have higher intervention success rates, suggesting that these models rely on their underlying cube state representation more.
The distributional-steering counterpart---change in total probability mass on the set of $\mathcal{T}$-good moves (post- minus pre-intervention)---is reported in Appendix~\ref{sec:appx_results}, Figure~\ref{fig:intervene_prob_mass}.

%% file: sections/4_grpo.tex
\section{RQ2: Effect of World Model on Post-Training}
\label{sec:planning}

\begin{figure*}[t]
\centering
\begin{minipage}[t]{0.497\textwidth}\centering
  {\footnotesize\textbf{\textsf{2$\times$2$\times$2}}}\\[2pt]
  \includegraphics[width=\textwidth]{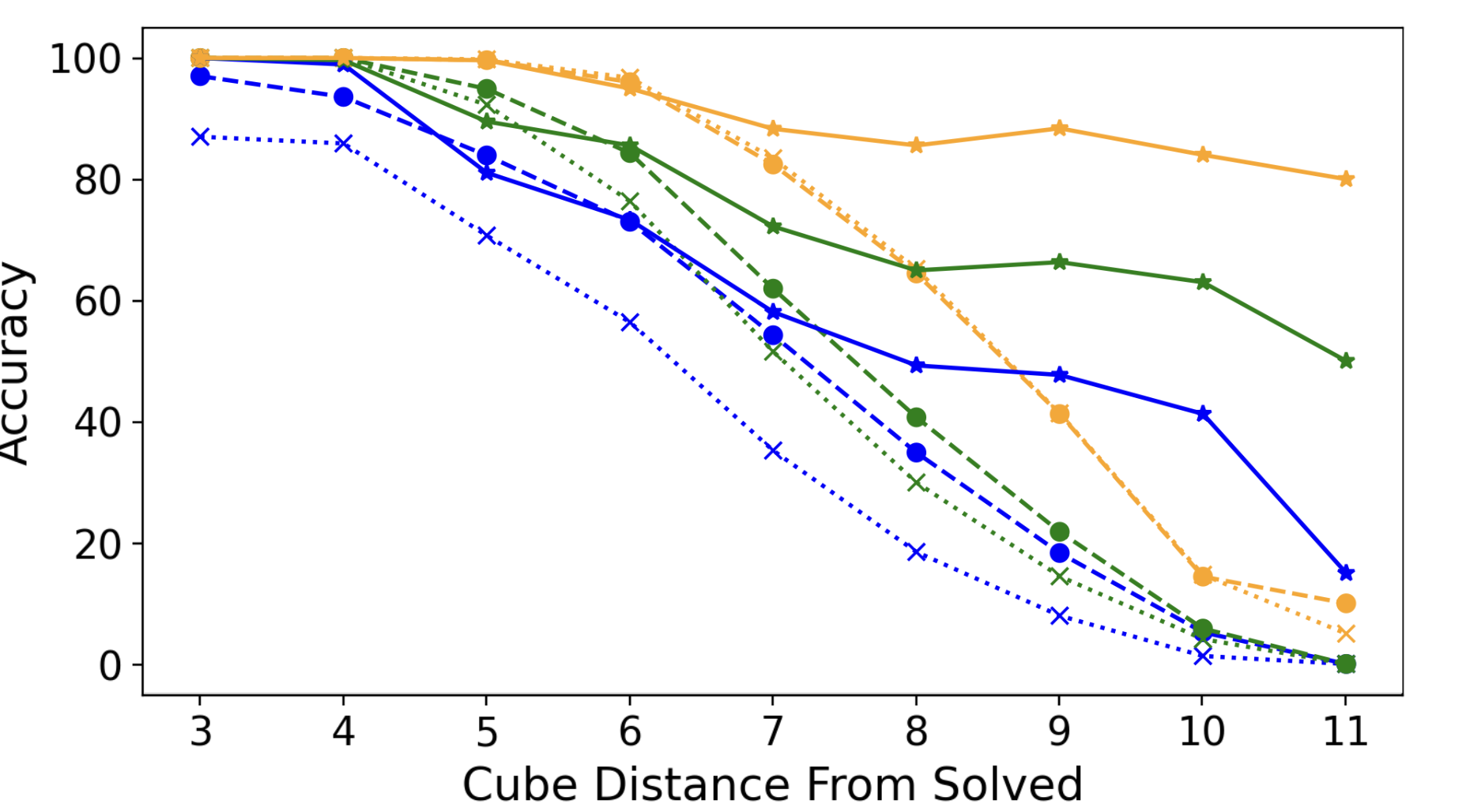}
\end{minipage}\hfill
\begin{minipage}[t]{0.497\textwidth}\centering
  {\footnotesize\textbf{\textsf{3$\times$3$\times$3}}}\\[2pt]
  \includegraphics[width=\textwidth]{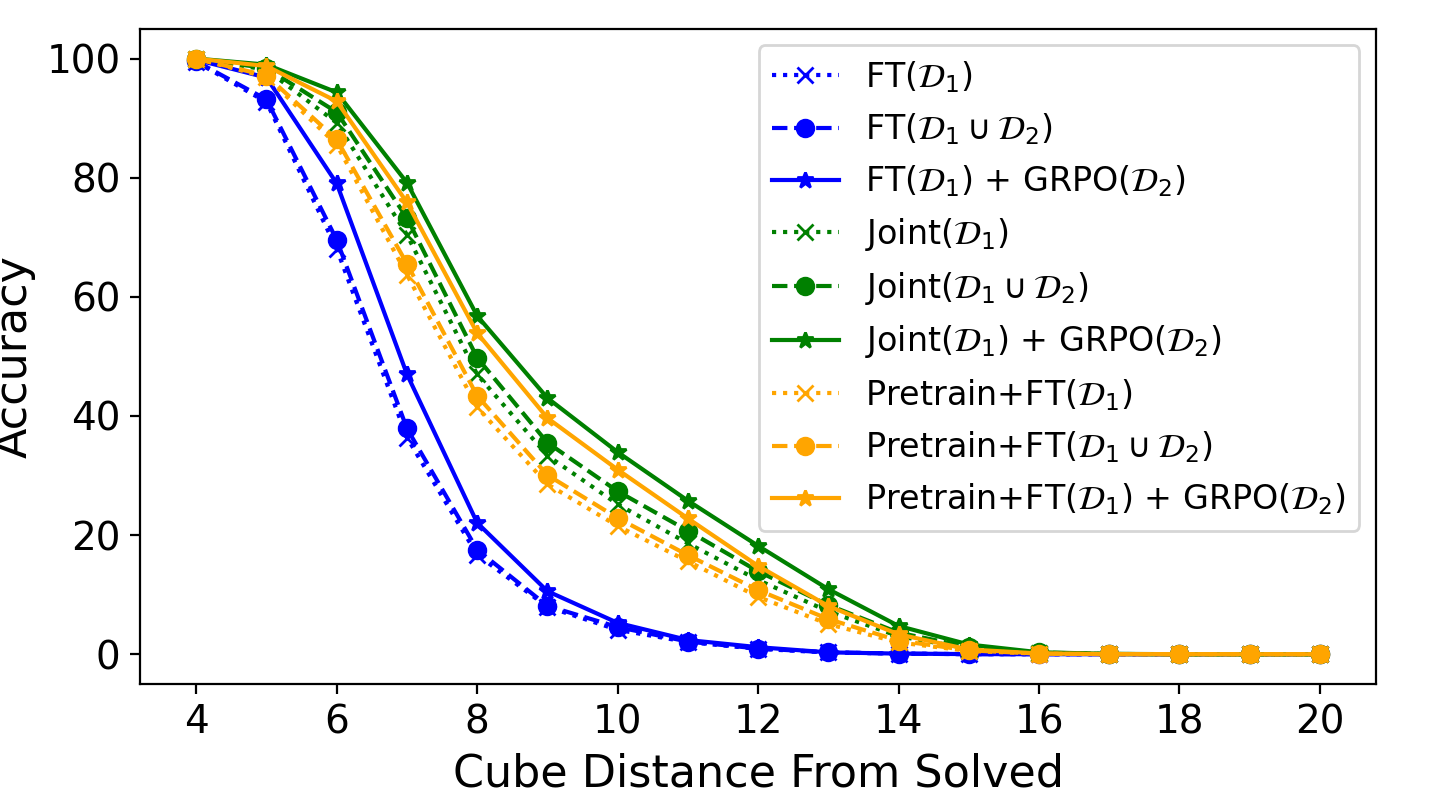}
\end{minipage}
\caption{\label{fig:grpo_results_222}\textbf{Task accuracy after GRPO, mean over 5 seeds.}
\emph{Left:} 2$\times$2. \emph{Right:} 3$\times$3.
Accuracy is broken down by cube distance. Dotted lines are $D_1$-only, dashed $D_1{\cup}D_2$, solid $D_1${+}GRPO.
On both cube sizes, applying GRPO instead of continued training improves accuracy, especially for harder cube states, and models explicitly trained on world modeling (Pretrain+FT, Joint) see the larger gains.}
\end{figure*}
Having established that explicitly training a world model improves cube state representations, we ask: what is the relationship between the quality of a world model and post-training?
We answer this first through task-accuracy comparisons of the configurations in Table~\ref{tab:training_setups} (Section~\ref{subsec:grpo_accuracy}), and then by quantitatively correlating probe accuracy with GRPO gain across the data-allocation models of both cube sizes (Section~\ref{subsec:repr_grpo}).

\subsection{Task Accuracy After GRPO}
\label{subsec:grpo_accuracy}

We measure task accuracy for 3 training strategies: standard fine-tuning (\textcolor{blue}{FT}), pretraining then fine-tuning (\textcolor{orange}{Pretrain+FT}), and joint-training (\textcolor{teal}{Joint}).
For each strategy we study 3 variants: applying it to (1) the first half of the data $D_1$, (2) the full data $D_1 \cup D_2$, or (3) $D_1$ followed by GRPO on $D_2$.

This yields 9 models for each of the 2$\times$2 and 3$\times$3 setup.
Figure~\ref{fig:grpo_results_222} reports the task accuracy of our models, defined as the percentage of scrambles for which the model generates a valid sequence that solves the initial state within $n{+}3$ moves ($n$ = optimal solution length), broken down by cube distance.
Each curve shows the average score from 5 seeds.
Within each color, comparing dashed lines (e.g., \textcolor{blue}{FT($D_1 \cup D_2$)}) against solid lines (\textcolor{blue}{FT($D_1$)+GRPO($D_2$)}) demonstrates the benefit of GRPO over continued training.
Comparing across colors shows the difference between training strategies.
For both the 2$\times$2 and 3$\times$3 case, we find that explicitly training a world model through pretraining or joint-training significantly outperforms FT, and that applying GRPO outperforms additional fine-tuning.

%\begin{table}[t]
%\centering
%\small
%\begin{tabular}{lccc}
%\toprule
%\textbf{Strategy} & $D_1$ & $D_1{\cup}D_2$ & $D_1${+}GRPO \\
%\midrule
%\textcolor{blue}{Fine-Tune}      & $14.7$ & $15.0$ & $16.9$ \\
%\textcolor{orange}{Pre-Train+FT}   & $23.5$ & $24.2$ & $27.9$ \\
%\textcolor{teal}{Joint Train}    & $25.7$ & $26.8$ & $\mathbf{29.6}$ \\
%\bottomrule
%\end{tabular}
%\caption{\label{tab:grpo_results_333}\textbf{Task accuracy after GRPO (3$\times$3$\times$3),} overall accuracy (\%) on the full distance-1--20 test set, mean over 5 seeds (standard deviations $\leq 1.5$, see Appendix~\ref{sec:appx_results}).
%GRPO is 3 epochs on $D_2$. Both explicit world-modeling strategies far exceed plain Fine-Tune, and GRPO lifts every strategy above its $D_1{\cup}D_2$ reference.}
%\end{table}

\subsection{Representation Quality vs. GRPO Gains}
\label{subsec:repr_grpo}

We further investigate the relationship between world-model quality and GRPO gains by holding fine-tuning and GRPO fixed while varying the amount of world-model pretraining.
For each resulting model we correlate its probe accuracy (mean over the last four layers and all timesteps), a proxy for world-model quality, with its GRPO gain (post-GRPO minus post-FT task accuracy).

\begin{figure*}[t]
\centering
\includegraphics[width=0.46\textwidth]{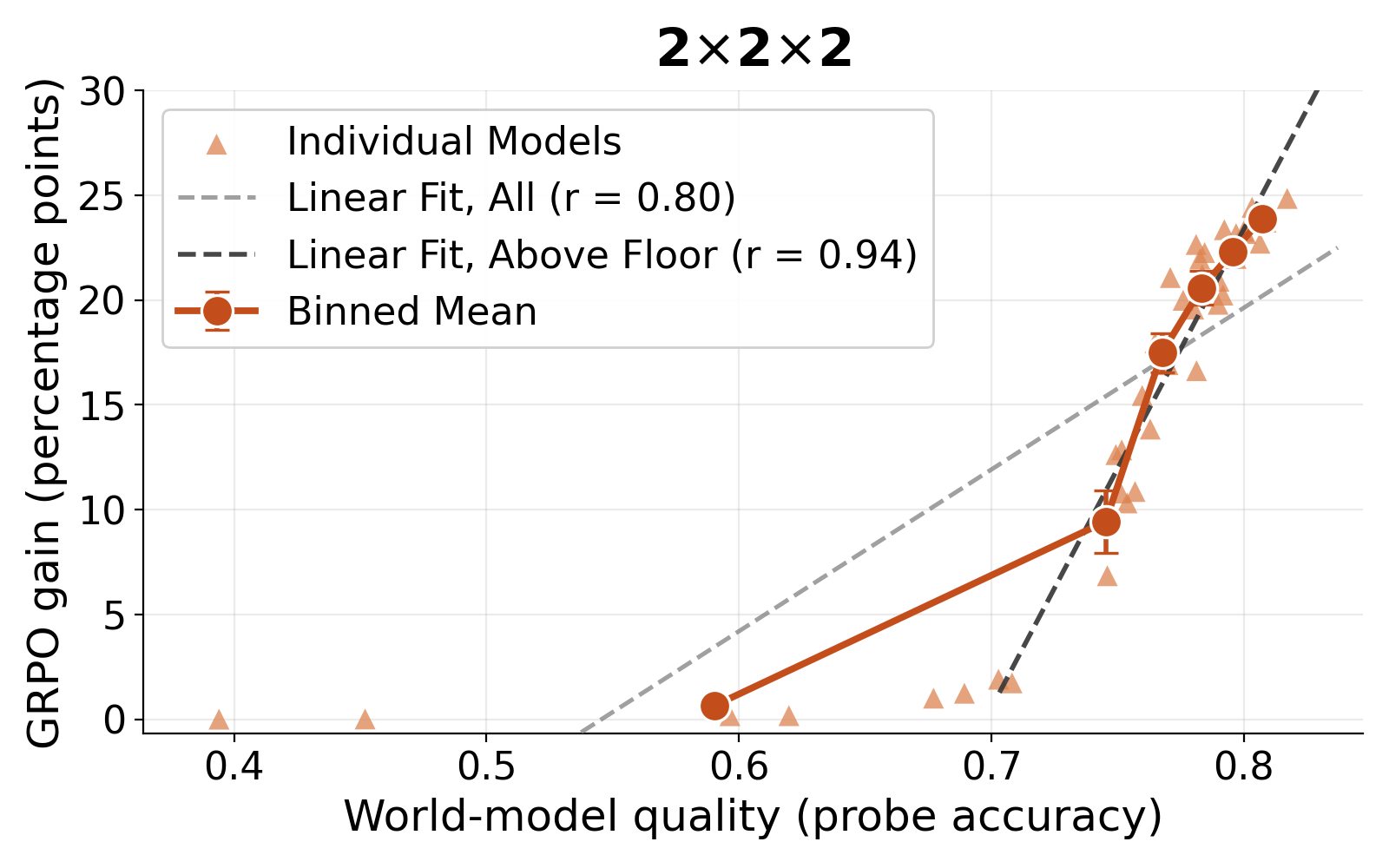}
\hfill
\includegraphics[width=0.46\textwidth]{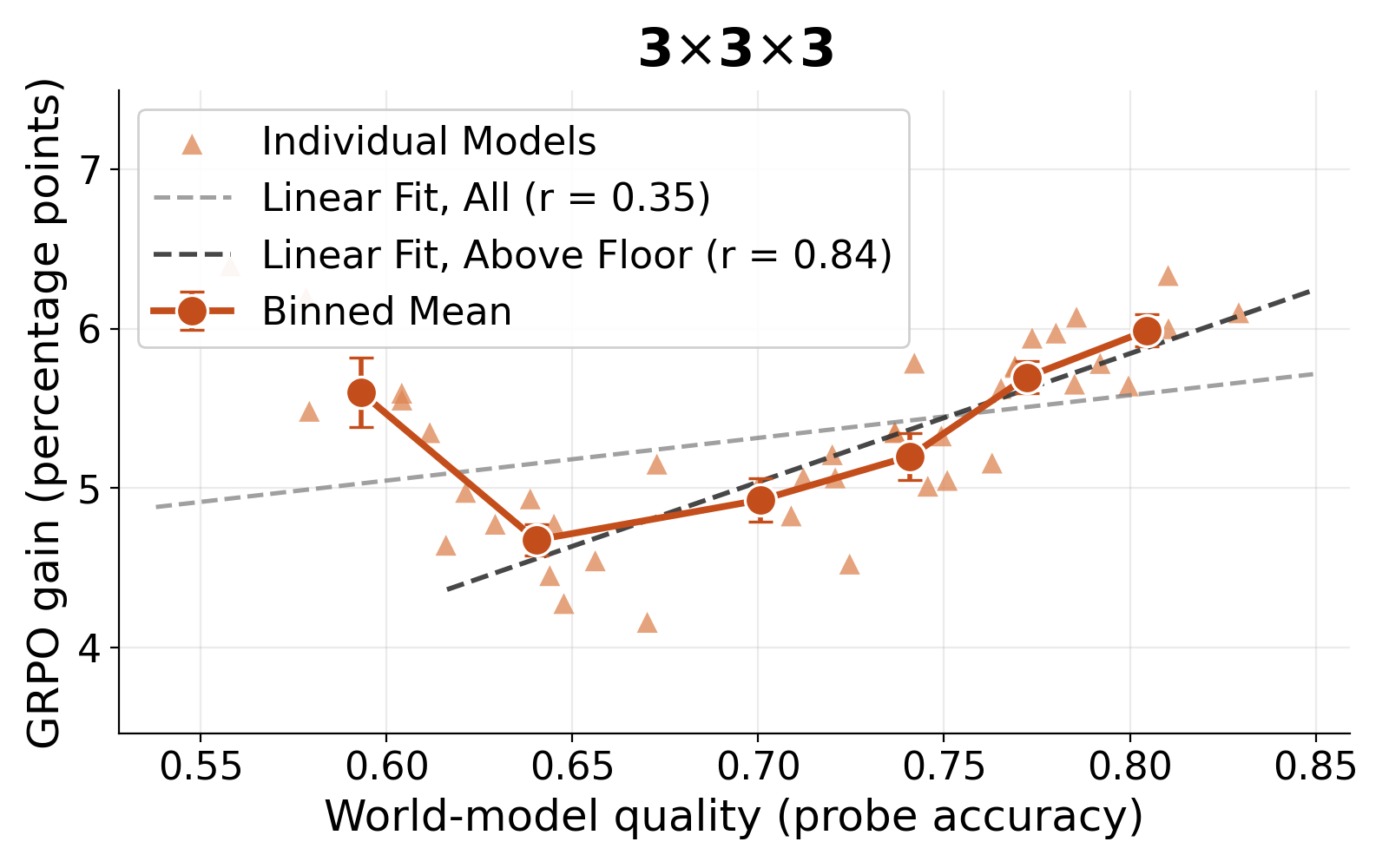}
\caption{\label{fig:probe_grpo_scatter}\textbf{Better world-model quality predicts larger GRPO gains.}
World-model quality (probe accuracy) versus GRPO gain across the 41 pretraining-amount configurations, which hold fine-tuning fixed and vary only pretraining.
Faint markers are individual models, the orange curve is the mean GRPO gain over six equal-count probe-accuracy bins ($\pm$SEM); the lighter dashed line is the linear fit through all 41 models, and the darker dashed line is the fit excluding the lowest bin---a degenerate floor regime detailed in Appendix~\ref{subsec:appx_bins} that flattens the all-configuration fit.
Above the floor, the relationship is clean and monotone on both cube sizes.}
\end{figure*}

Figure~\ref{fig:probe_grpo_scatter} demonstrates a positive correlation between world-model quality and GRPO gain on both cube sizes when fitting all 41 configurations (Pearson $r{=}0.80$ on 2$\times$2, $r{=}0.35$ on 3$\times$3).
A single linear fit through all 41 models understates the trend, however, because a handful of barely-pretrained models behave atypically: on 2$\times$2 their world model is too weak for GRPO to find any reward to amplify, while on 3$\times$3 they solve only a small fraction of easy cubes and their raw GRPO gain reflects sheer headroom from a low base rather than world-model quality.
Dropping the lowest equal-count probe-accuracy bin (the darker ``Above Floor'' fit, $N{=}34$) tightens the correlation sharply on both cubes ($r{=}0.94$ on 2$\times$2, $r{=}0.84$ on 3$\times$3); we report the bin-level diagnostics in Appendix~\ref{subsec:appx_bins}.
Above the floor, every step up in world-model quality is associated with larger GRPO gains; on 2$\times$2 the gains rise steeply once probe accuracy passes around $0.75$.

World-model quality also predicts where a model ends up, not just how far GRPO moves it: on 3$\times$3, probe accuracy tracks absolute task accuracy nearly rank-perfectly, both before and after GRPO (Pearson $r{=}0.96$ post-SFT, $r{=}0.98$ post-GRPO; Appendix~\ref{subsec:appx_bins}).
A model that builds a better internal cube is the better solver both before and after RL.

%% file: sections/5_data_allocation.tex
\section{RQ3: Data Allocation for World Modeling}
\label{sec:data_allocation}

Thus far we have demonstrated that explicitly learning a world model leads to better post-training performance.
Here we consider a more realistic scenario observed in practice: given a fixed computational budget, how much should be spent on explicit world-modeling?

We consider two settings.
First, we allocate a small, fixed budget for fine-tuning and GRPO while controlling for pretraining (Section~\ref{subsec:fixed_sft}).
This allows us to study the impact of additional world-model pretraining on downstream training, as well as how far we can get with minimal post-training on top of world models.
Second, we keep our GRPO budget fixed and let pretraining and fine-tuning compete for one budget (Section~\ref{subsec:ratio_sweep}), simulating a situation a practitioner actually faces.

For the 3$\times$3 experiments we restrict to the strict $\leq 11$ subset of the data, matching the 2$\times$2 maximum distance for a direct cross-size comparison.

\subsection{How far can we get with explicit world-modeling and minimal post-training?}
\label{subsec:fixed_sft}

Here we are interested in studying how much explicit world-modeling can provide with minimal fine-tuning and GRPO.
Thus we hold the fine-tuning and GRPO stages fixed and only vary the pretraining budget.
We purposefully allocate a small, fixed training budget for SFT and GRPO (3 epochs for SFT on a fixed slice of $D_1$, 1 GRPO epoch on $D_2$).
Note that this is a much smaller budget compared to our results from Section~\ref{sec:rq1_pretrain} and \ref{sec:planning}, which use SFT with early stopping (no fixed epoch count) and 3 epochs of GRPO.
The pretraining data comes from a second, separate slice of $D_1$ that fine-tuning never sees, so adding more pretraining never costs us any SFT data.
We sweep both how much of that pretrain pool the model gets (in eighths, from none to all) and how many passes it makes over it (1 to 5 epochs), resulting in 41 models per cube size.

\begin{figure*}[t]
\centering
\includegraphics[width=0.49\textwidth]{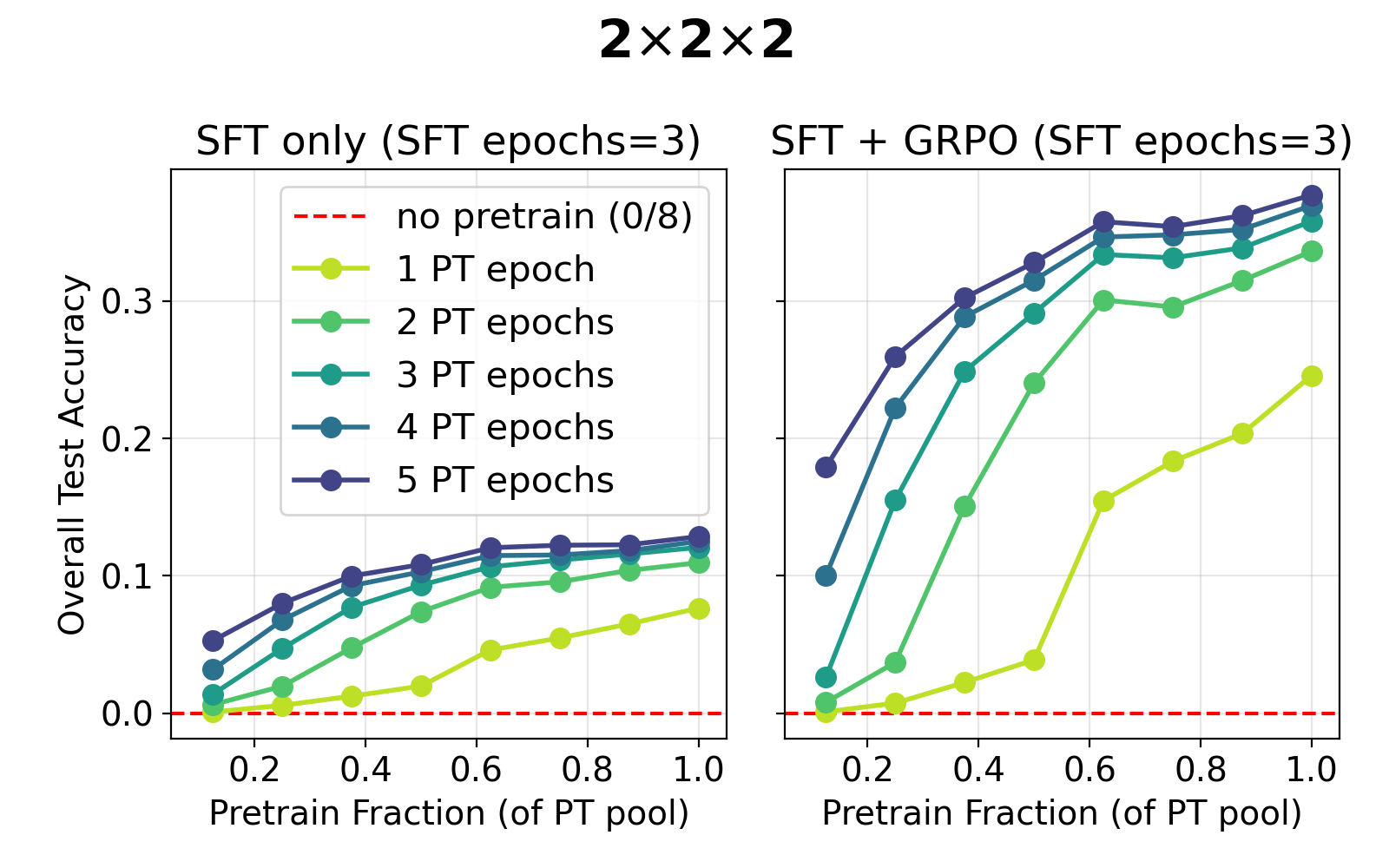}
\hfill
\includegraphics[width=0.49\textwidth]{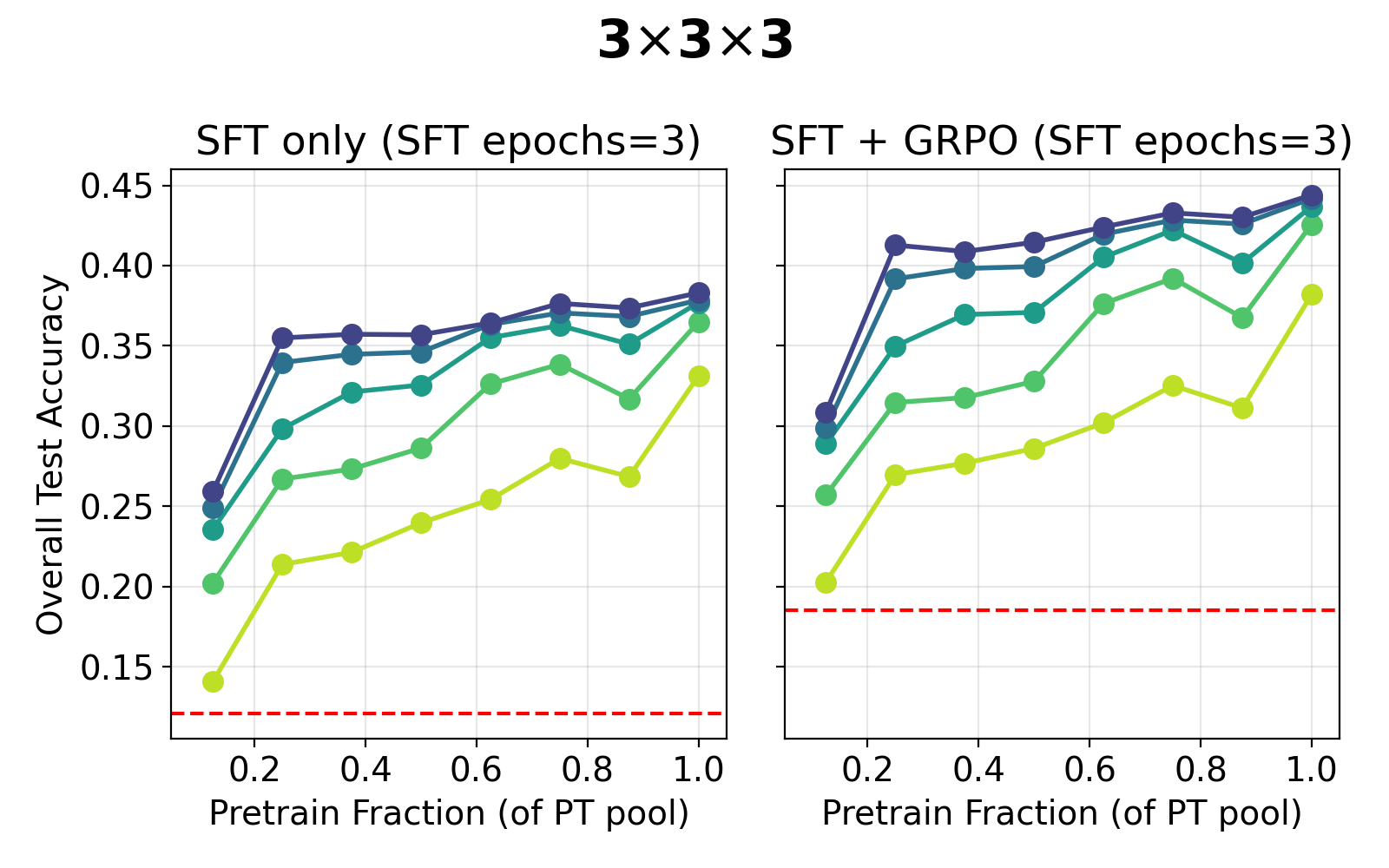}
\caption{\label{fig:fxsft_acc_222}\textbf{Accuracy vs.\ pretrain-pool fraction.}
\emph{Left:} 2$\times$2$\times$2. \emph{Right:} 3$\times$3$\times$3 ($\leq 11$ subset).
Each half shows SFT-only and SFT+GRPO accuracy. Each curve is a pretraining-epoch count, with the dashed line the no-pretrain ($k{=}0$) baseline.
Accuracy rises monotonically with pretraining on both cube sizes: with the SFT budget held fixed, more pretraining only helps.}
\end{figure*}

\begin{figure*}[t]
\centering
\includegraphics[width=0.44\textwidth]{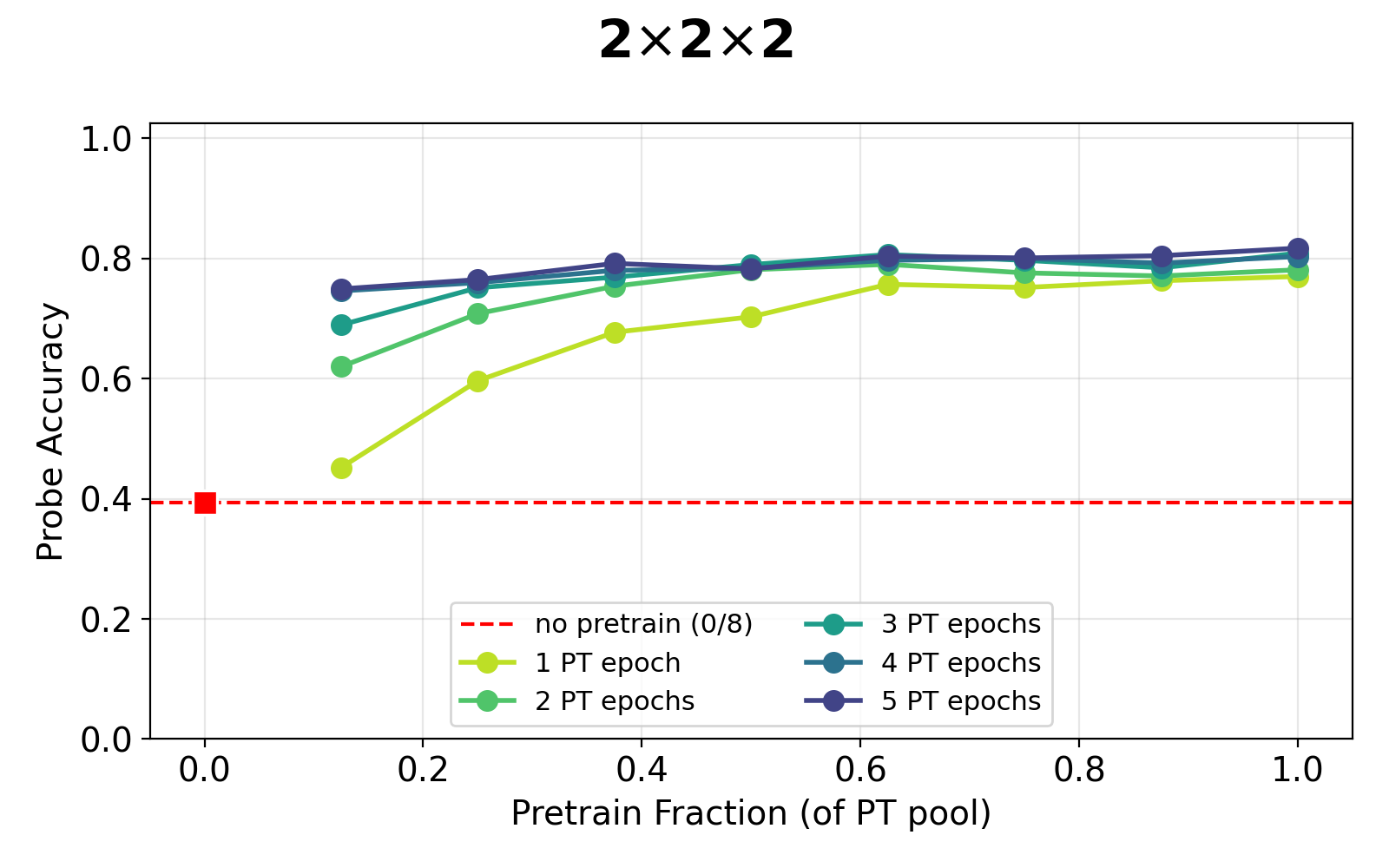}
\hfill
\includegraphics[width=0.44\textwidth]{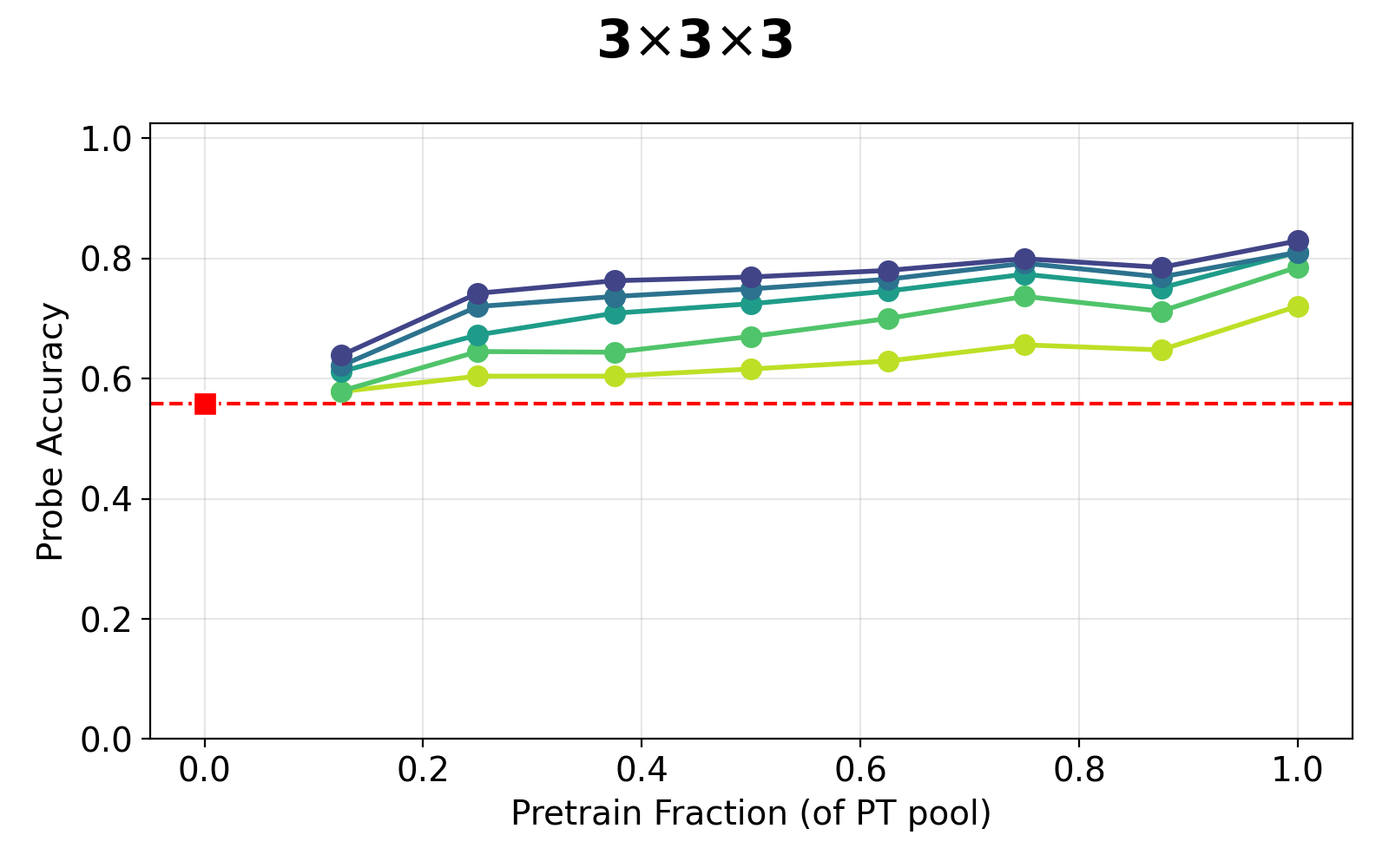}
\caption{\label{fig:fxsft_saturation}\textbf{Probe accuracy vs.\ pretrain-pool fraction.}
\emph{Left:} 2$\times$2$\times$2. \emph{Right:} 3$\times$3$\times$3 ($\leq 11$ subset).
Mean probe accuracy (last 4 layers, all timesteps) vs.\ pretrain-pool fraction, one curve per pretraining-epoch count.
Probe accuracy rises monotonically with pretraining and saturates quickly: on 2$\times$2$\times$2 the no-pretrain baseline is $0.39$ and $1/8$ of the pool at 5 epochs already reaches $0.75$. On 3$\times$3$\times$3 the no-pretrain baseline is higher ($0.56$), but the same rising, saturating shape holds.}
\end{figure*}

Figure~\ref{fig:fxsft_acc_222} demonstrates a clear trend.
With fine-tuning and GRPO held fixed, more pretraining gives us a better model both after SFT and GRPO.
This in and of itself is not surprising, as more pretraining is known to lead to better models.
More importantly, we are able to observe how far we can get with explicit world-modeling and minimal post-training.
On 2$\times$2, no pretraining leaves both SFT and SFT+GRPO accuracy near zero, while the best run reaches $12.9\%$ after SFT and $37.7\%$ after GRPO.
The benefits are especially noticeable for GRPO: presumably, explicit world-modeling provides GRPO the right features (cube representations) for GRPO to learn the task.
The same trend can be found for the 3$\times$3 cube, albeit the drastic effect on GRPO is less pronounced.
The two accuracy curves track probe accuracy (Figure~\ref{fig:fxsft_saturation}): both SFT and GRPO inherit the world model that pretraining builds.

\subsection{Splitting a Fixed Budget}
\label{subsec:ratio_sweep}

Given a fixed training-data budget, how much of it should be spent on explicit world-modeling versus fine-tuning?
We study this question by splitting a single pool ($D_1$) between pretraining and fine-tuning and sweep over how it is split, from pure fine-tuning to mostly pretraining, in eighths.
Each split is trained at five training lengths $N \in \{1,\dots,5\}$, where $N$ sets the number of epochs for both the pretraining and SFT stages, with GRPO unchanged.

\begin{figure*}[t]
\centering
\includegraphics[width=0.49\textwidth]{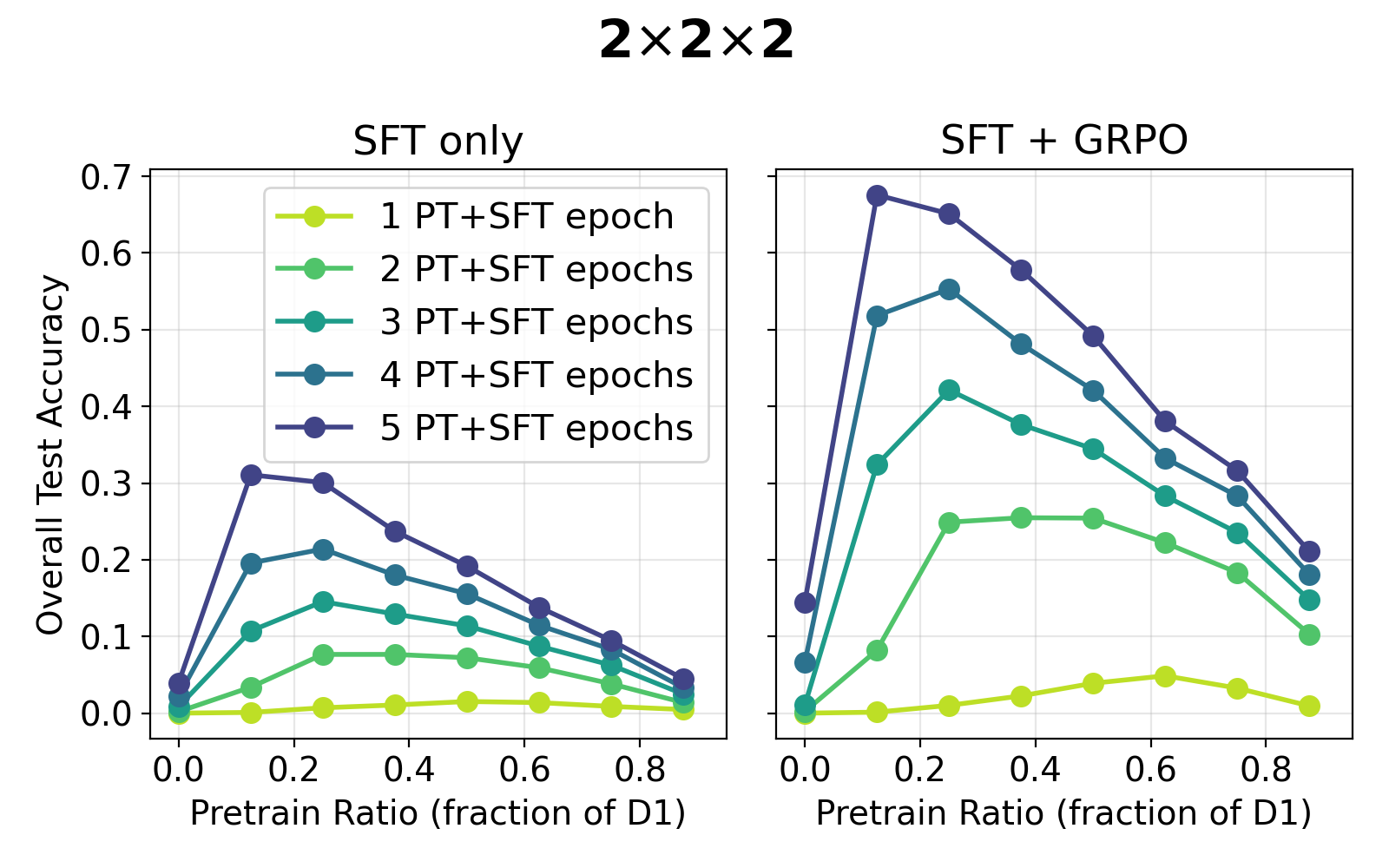}
\hfill
\includegraphics[width=0.49\textwidth]{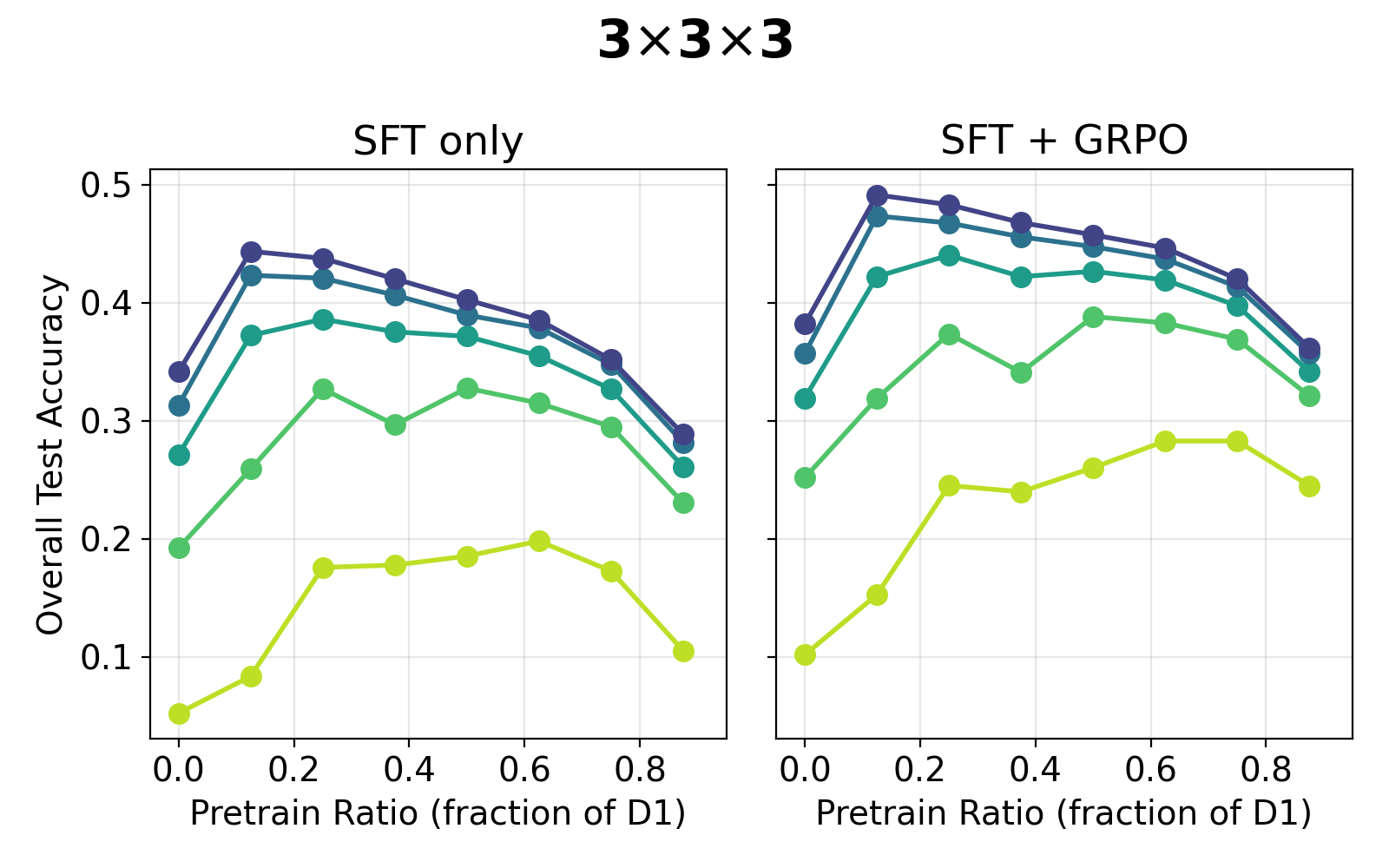}
\caption{\label{fig:sweep_222}\textbf{Accuracy vs.\ pretrain:SFT ratio.}
\emph{Left:} 2$\times$2$\times$2. \emph{Right:} 3$\times$3$\times$3 ($\leq 11$ subset).
Each half shows SFT-only and SFT+GRPO accuracy. Each curve is a training length $N \in \{1,\dots,5\}$, where $N$ is the number of epochs run for both pretraining and SFT.
On both cube sizes accuracy peaks at $k{=}1/8$ at $N{=}5$, and the peak drifts leftward (toward less pretraining) as $N$ grows.}
\end{figure*}

Figure~\ref{fig:sweep_222} shows our results.
There are a few takeaways worth noting.
First, a little pretraining goes a long way.
Accuracy peaks sharply when allocating just an eighth of the budget on pretraining, then falls as further pretraining starves fine-tuning, suggesting that additional pretraining is marginal and better spent on fine-tuning.
This pattern holds for both cube sizes and for most values of $N$.

Put together, our results indicate that there is a sweet-spot regarding allocating between explicit world-modeling, fine-tuning, and post-training with GRPO.
While our findings pertain to our specific Rubik's Cube setup, we speculate that similar findings may hold for different domains, in that even a little bit of explicit world-modeling can lead to large benefits for downstream training stages.

%% file: sections/6_related_work.tex
\section{Related Work}
\label{sec:related_work}

\paragraph{World modeling.}
We use the term ``world model'' to refer to a model's ability to track the state of its (latent) environment and understand how its actions affect that state~\citep{ha2018world, li2025does}.
\citet{li2023emergent} show that Transformers trained on game-move transcripts can reconstruct the latent board state. \citet{nanda2023emergent} found these representations to be linearly decodable, and \citet{karvonen2024emergent} extended the result to chess.
Unlike these works, which study the emergence of world models, we study how explicit world-model objectives interact with downstream training stages.

\paragraph{Probing and causal interventions.}
A growing line of work relies on linear probes to decode interpretable representations from a model's hidden states~\citep{alain2016understanding}, uncovering that many human-interpretable concepts are linearly encoded in model activations~\citep{lee2024mechanistic, lee2025shared, park2023linear, park2024geometry, gurnee2023language, burns2023discovering}.
A common practice is to verify the role of such representations through causal interventions~\citep{arditi2024refusal}.

\paragraph{Post-training with reinforcement learning.}
Reinforcement learning from human feedback and its variants are now standard for post-training language models~\citep{ouyang2022training}.
GRPO~\citep{shao2024deepseekmath} uses group-relative rewards without a separate critic and has been applied at scale in reasoning-focused training~\citep{guo2025deepseek}.
We instead study how the quality of the representations entering the RL stage affects the RL outcome, which to our knowledge has not been examined directly in controlled settings. Related model-based RL work uses learned world models to generate imagined experience for policy learning~\citep{young2023benefitsmodelbasedgeneralizationreinforcement}, while our state-prediction heads are used only as training objectives.

\paragraph{Data allocation and curriculum learning.}
How to allocate data across training stages relates to curriculum learning~\citep{bengio2009curriculum} and domain-specific pretraining~\citep{azerbayev2024llemma}.
Our data-allocation experiments provide controlled evidence for this question in a setting where world-model quality can be precisely measured alongside task accuracy.

%% file: sections/7_conclusion.tex
\section{Conclusion}
\label{sec:conclusion}

We study the relationship between explicit world-model training, internal representations, and post-training performance across two Rubik's Cube sizes.
First, explicitly pretraining a world model leads to more robust internal representations, evidenced by higher linear probe accuracies and greater causal steerability.
Second, models with these improved representations see larger GRPO gains, especially on harder cube states. When only the amount of pretraining is varied, probe accuracy strongly predicts the GRPO gain, a direct quantitative relationship between representation quality and GRPO improvement.
Third, under a fixed total budget, accuracy is maximized by allocating only a small fraction (roughly an eighth) to pretraining, with the optimum drifting toward more pretraining when training is short.
The probing, GRPO, and data-allocation results all hold across both cube sizes.
These results motivate testing the same approach in other sequence-planning tasks where intermediate states are available, such as program execution, theorem proving, games, and simulators. If the same patterns hold in such tasks, investing in world-model quality before policy learning may improve both supervised and reinforcement-learning post-training, and the data split between these stages may deserve careful tuning. Code is available at \url{https://github.com/prakharg55/CubeLM-EMNLP}.